\documentclass[preprint,12pt]{elsarticle}

\usepackage[utf8]{inputenc} 
\usepackage[T1]{fontenc}    
\usepackage{hyperref}       
\usepackage{url}            
\usepackage{booktabs}       
\usepackage{amsfonts}       
\usepackage{nicefrac}       
\usepackage{microtype}      
\usepackage{xcolor}
\usepackage{algorithm}
\usepackage{algorithmic}
\usepackage{mathtools}
\usepackage{booktabs}
\usepackage{longtable}
\usepackage{array}
\usepackage{multirow}
\usepackage{wrapfig}
\usepackage{float}
\usepackage{colortbl}
\usepackage{pdflscape}
\usepackage{tabu}
\usepackage{threeparttable}
\usepackage{threeparttablex}
\usepackage[normalem]{ulem}
\usepackage{makecell}
\usepackage{xcolor}
\usepackage{adjustbox}
\usepackage{multicol}
\usepackage{caption}
\usepackage{subcaption}
\usepackage{todonotes}
\usepackage{svg}

\journal{Knowledge-Based Systems}

\begin{document}

\begin{frontmatter}
\title{An Explainable Multi-Task Similarity Measure: Integrating Accumulated Local Effects and Weighted Fr\'echet Distance}

\author[aff1,aff2]{Pablo~Hidalgo\corref{cor1}}
\author[aff1]{Daniel~Rodriguez}

\cortext[cor1]{Corresponding author}

\affiliation[aff1]{organization={Department of Computer Science, University of Alcala},
postcode={28805},
city={Alcal\'a de Henares, Madrid},
country={Spain}}

\affiliation[aff2]{organization={Faculty of Law, Business and Government, Universidad Francisco de Vitoria},
addressline={Ctra. Pozuelo-Majadahonda Km 1,800},
postcode={28223},
city={Pozuelo de Alarc\'on, Madrid},
country={Spain}}

\begin{abstract}
In many machine learning contexts, tasks are often treated as interconnected components with the goal of leveraging knowledge transfer between them, which is the central aim of Multi-Task Learning (MTL). Consequently, this multi-task scenario requires addressing critical questions: which tasks are similar, and how and why do they exhibit similarity? In this work, we propose a multi-task similarity measure based on Explainable Artificial Intelligence (XAI) techniques, specifically Accumulated Local Effects (ALE) curves.

ALE curves are compared using the Fréchet distance, weighted by the data distribution, and the resulting similarity measure incorporates the importance of each feature. The measure is applicable in both single-task learning scenarios, where each task is trained separately, and multi-task learning scenarios, where all tasks are learned simultaneously. The measure is model-agnostic, allowing the use of different machine learning models across tasks. A scaling factor is introduced to account for differences in predictive performance across tasks, and several recommendations are provided for applying the measure in complex scenarios.

We validate this measure using four datasets, one synthetic dataset and three real-world datasets. The real-world datasets include a well-known Parkinson’s dataset and a bike-sharing usage dataset --- both structured in tabular format --- as well as the CelebA dataset, which is used to evaluate the application of concept bottleneck encoders in a multitask learning setting. The results demonstrate that the measure aligns with intuitive expectations of task similarity across both tabular and non-tabular data, making it a valuable tool for exploring relationships between tasks and supporting informed decision-making.

\end{abstract}


\end{frontmatter}

\section{Introduction}
\label{sec:intro}

In a complex world, addressing tasks as interconnected elements rather than isolated parts makes intuitive sense, especially in a multi-task scenario. This paper specifically focuses on artificial intelligence tasks that can be learned by machine learning models from two perspectives: (i) in a single-task manner, where each task is learned independently by its own model, which may vary across tasks, or (ii) in a multi-task manner, where tasks are learned jointly, leveraging shared information and transferring knowledge between tasks to achieve better results~\cite{caruana_multitask_1997,zhang_survey_2022}.

Whether employing single-task or multi-task learning, some critical questions arise depending on the context: (i) which tasks are similar, (ii) to what extent are they similar, and (iii) why are they similar?

Identifying relationships between tasks is highly relevant in various contexts, including cybersecurity~\cite{zhang_explainable_2022}, manufacturing~\cite{chen_explainable_2023}, and healthcare~\cite{chaddad_survey_2023}. Understanding task similarities is crucial for determining whether similar tasks exhibit analogous behaviors in actionable contexts. For example, two diseases can be treated as distinct tasks, but their similarity might suggest that they respond similarly to a specific treatment. To leverage such relationships, it is essential not only to identify similar tasks but also to explain the basis of their similarity.

Current multi-task learning techniques compute task similarity in an abstract manner, which may introduce biases in the learning algorithm. For instance, soft parameter-sharing techniques in deep learning impose penalties on the loss function using metrics such as the Frobenius norm~\cite{ruder_overview_2017}. While these approaches often achieve more accurate models in terms of loss minimization, they lack interpretability regarding how and why tasks are similar.

Explainable Artificial Intelligence (XAI)~\cite{barredo_arrieta_explainable_2020} has been a critical tool in understanding the inner workings of black-box models, providing transparency in predictions and enhancing trust. However, to the best of our knowledge, current explainable techniques operate only in single-task scenarios, focusing solely on understanding how a single model works~\cite{dosilovic_explainable_2018}.

The aim of this paper is to propose a \textit{multi-task similarity measure to identify and explain task similarities in an interpretable manner}. The computation of this measure is agnostic to whether tasks were trained using single-task or multi-task learning; it only requires the trained models for each task. Importantly, the measure evaluates the similarities between the models of the tasks, not directly on the datasets. This approach enables us to analyze the mechanisms of the models and their relationships. Since models are simplifications of the underlying complexities of tasks, ensuring sufficient model quality is crucial.

This work makes the following contributions to achieve this objective: 

\begin{itemize} 
\item We define a similarity measure between tasks from an interpretability perspective.
\item The similarity measure introduces weights that account for the importance of each variable in a task and the reliability of each segment of the data.
\end{itemize}

In this work, the chosen interpretability method is Accumulated Local Effects (ALE) curves~\cite{apley_visualizing_2020}. ALE curves capture the average influence of each feature on predictions and offer advantages over other explainability tools, such as Partial Dependence Plots (PDPs)~\cite{friedman_greedy_2001}. However, other methods can also be easily incorporated into the measure.

To quantify the similarity between tasks, we develop a weighted modification of the Fr\'echet Distance~\cite{eiter_computing_1994}, a standard measure for computing the distance between curves. Furthermore, we propose a scaling factor to modify the similarity to account for differences in predictive performance across task, and several recommendations are provided for applying the measure in complex scenarios such as dealing with non-tabular data (e.g., images) or heterogeneous feature spaces across tasks. These guidelines help ensure the robustness and interpretability of the similarity measure under a wide range of real-world conditions.

The rest of the paper is organized as follows. Section~\ref{sec:background} reviews the background and related work, covering key concepts such as multi-task learning, explainable artificial intelligence, and the Fr\'echet distance. In Section~\ref{sec:similarity}, we provide the formal definition of the proposed multi-task similarity measure. Section~\ref{sec:empiricalWork} applies the measure to four datasets: a synthetic dataset, a Parkinson’s patient dataset, a bike-sharing usage dataset, and an image dataset (CelebA). Section~\ref{sec:results-discussion} discusses the main properties, strengths, and weaknesses of the measure. Finally, Section~\ref{sec:conclusions} presents the conclusions and outlines future directions for this work.

\section{Background and related work}
\label{sec:background}

\subsection{Multi-task Learning}
\label{subsec:mtl}

The Multitask Learning (MTL) paradigm~\cite{caruana_multitask_1997} is an active research area that aims to leverage shared knowledge between tasks to alleviate data sparsity and reduce learning errors. The main hypothesis of MTL is that simultaneously learning multiple related tasks can enhance the generalization performance across all considered tasks~\cite{zhang_survey_2022}. A critical aspect involves determining which tasks are similar to effectively transfer knowledge between them. Otherwise, multi-task learning may not function optimally, as knowledge transfer from unrelated tasks can detrimentally impact learning and worsen single-task learning. 

State-of-the-art techniques in MTL can be broadly categorized into hard parameter sharing, soft parameter sharing, and task-specific modules~\cite{ruder_overview_2017}.

Hard parameter sharing approaches, such as those used in early neural networks, involve sharing a common set of parameters between all tasks, thereby reducing the risk of overfitting but limiting task-specific flexibility.

In contrast, soft parameter sharing techniques, such as those based on matrix factorization or attention mechanisms, maintain task-specific parameters while imposing regularization constraints to encourage similarity~\cite{zhang_survey_2022}. Recent advances have introduced novel architectures such as the multigate mixture-of-experts (MMoE), which dynamically allocate resources to tasks and achieve state-of-the-art results in applications ranging from natural language processing to computer vision~\cite{ma_modeling_2018}. Transformer-based models, which leverage attention mechanisms to model inter-task relationships, have also demonstrated superior performance in multi-task learning~\cite{yu_unleashing_2024}.

Despite their success, a persistent challenge in MTL is understanding and quantifying task relationships, as most methods rely on abstract or implicit measures of similarity, limiting interpretability.

\subsection{Explainable AI}
\label{subsec:xai}

The field of Explainable Artificial Intelligence (xAI)~\cite{dosilovic_explainable_2018} has gained substantial attention recently as the deployment of complex machine learning models in real-world applications has become more prevalent in domains such as the medical domain, where explainability is a must. Understanding and interpreting the decisions made by these complex models is crucial for building trust, ensuring accountability, and meeting ethical standards. This requires the development of techniques capable of explaining the diverse range of models employed in various contexts. Furthermore, explainability can be a crucial tool in the refinement of the process of developing a machine learning model~\cite{yin_explainable_2022}.

There exist diverse types of explainable methods, each exploiting certain properties of the models. Local methods aim to explain individual predictions or a small subset of data such as Individual Conditional Expectation (ICE)~\cite{goldstein_peeking_2015}, Local Interpretable Model-agnostic Explanations (LIME)~\cite{ribeiro_why_2016} or SHapley Additive exPlanations (SHAP)~\cite{lundberg_unified_2017}. Global methods, however, try to explain the behavior of the entire model and provide a general picture of the overall trends. An important global model-agnostic method is the ALE plots~\cite{apley_visualizing_2020}, which allows for an understanding of how one or two features influence the prediction on average. This method is unbiased and less computationally expensive than other alternatives, such as PDP~\cite{friedman_greedy_2001}. In this work, the multi-task similarity measure is developed using ALE as the foundation to compute the similarity between features across different tasks.

However, current xAI techniques operate in a single-task manner and are designed to understand only the behavior of individual models and their variables. As a result, they do not effectively address the multi-task paradigm or facilitate an understanding of inter-task relations in order to explain the "big picture".

\subsection{Similarity measures}
\label{subsec:similarityMeasures}

A similarity measure can be defined as a function that assesses the degree of similarity or the relationship between two entities~\cite{seel_measures_2012}. In this work, the considered entities are tasks represented as mathematical models resulting from a machine learning process. There are many definitions of similarity, depending on the nature of the context at hand and the properties that we want the similarity to satisfy~\cite{manning_introduction_2008}. The Euclidean distance and cosine similarity are two classical measures that capture the relationship between vectors, and they have a broad application in many fields. In natural language processing and bioinformatics, edit distance and string kernels have been essential for measuring the similarity between sequences~\cite{riesen_approximate_2009,leslie_spectrum_2001}. In statistics, the Kullback-Leibler divergence~\cite{kullback_information_1951} is an important measure that works with two probability distributions, and it quantifies the differences between them.

In this work, each task is represented by its ALE curves for each variable. Therefore, the similarity between tasks must be computed based on these curves. In the literature, there are several measures to compute the similarity between curves. 

One important similarity measure between curves is the Hausdorff distance~\cite{alt_computing_1995} which for arbitrary bounded sets $A,B\subseteq \mathbb{R}^2$ is defined as:

\begin{equation*}
    \delta_{H}(A,B) := max\Big(\underset{a\in A}{sup}\,\underset{b\in B}{inf}d(a, b),\underset{b\in B}{sup}\,\underset{a\in A}{inf}d(a, b)\Big)
\end{equation*}

where $d$ is the Euclidean distance. In other words, the Hausdorff distance measures the maximum distance an adversary can force you to travel by selecting a point in one set, from which you must then travel to the nearest point in the other set. In simpler terms, it represents the greatest distance from a point in one set to the closest point in the other set. This distance has numerous applications, primarily in image comparison, such as in segmentation algorithms for medical images~\cite{morain-nicolier_hausdorff_2007} or in computer graphics~\cite{cignoni_metro_1998}. However, the Hausdorff distance only considers the sets of points along both curves and does not capture the trajectories of the curves themselves. In many applications, such as the focus of this paper, understanding the course of the ALE curves is crucial as they summarize the behavior of each variable in each task. 

The similarity measure proposed in this paper is based on the Fr\'echet distance~\cite{frechet_sur_1906}, which is a measure of similarity between curves that considers both the location and, importantly, the ordering of the points along the curves. If a curve is defined as a continuous mapping $f:[a,b]\rightarrow V$ where $a,b\in\mathbb{R}$ and $a\leq b$ and $(V,d)$ is a metric space then, given two curves $f:[a,b]\rightarrow V$ and $g:[a',b']\rightarrow V$, the Fr\'echet distance between them is defined~\cite{eiter_computing_1994},~\cite{alt_computing_1995} as

\begin{equation}
    \delta_{F}(f,g) = \underset{\alpha,\beta}{inf}\underset{t\in[0,1]}{max}d(f(\alpha(t),g(\beta(t)))
\end{equation}

where $\alpha$ and $\beta$ are both arbitrary continuous non-decreasing functions from $[0,1]$ onto $[a,b]$ and $[a',b']$, respectively. Usually, this distance is intuitively explained as follows: a person walks a dog on a leash, with the person following one curve, while the dog following another. Both are free to adjust their speed, but they cannot backtrack. The Fr\'echet distance represents the minimum length of the leash required for traversing both curves~\cite{eiter_computing_1994,alt_computing_1995}.

The exact Fr\'echet distance between two polygonal curves can be computed in time $\mathcal{O}(pqlog^2 pq)$~\cite{alt_computing_1995} where $p$ and $q$ are the number of segments on the polygonal curves.

However, if we only focus on the positions of the endpoints of the line segments defining the polygonal curves, we encounter the discrete Fr\'echet distance, also known as the coupling distance~\cite{eiter_computing_1994}. It can be proved that this distance serves as an upper bound for the (continuous) Fr\'echet distance, and the difference between them is bounded by the length of the longest edge of the polygonal curve~\cite{eiter_computing_1994}. As this distance forms the basis for the similarity measure defined here, the formal definition of the discrete Fr\'echet distance is discussed in Section~\ref{subsec:wFrechet}. An algorithm based on dynamic programming can compute the discrete Fr\'echet distance in time $\mathcal{O}(pq)$~\cite{eiter_computing_1994}. Furthermore, in this work it is convenient to use a variant of the discrete Fr\'echet distance that uses the sum instead of the maximum as exposed by Eiter and Mannila~\cite{eiter_computing_1994}.

In our approach, the tasks are simplified and represented as curves, specifically ALE curves, from an explainable standpoint. With a slight adaptation, the discrete Fr\'echet distance serves as a suitable foundation for calculating the similarity between these curves. Note that the definition of the similarity concept developed here assumes that lower values imply high similarity and vice versa (in other works, this is referred to as dissimilarity).

In this paper, we propose a post-hoc multi-task similarity measure, assuming that tasks have been trained using a single or multi-task learning paradigm. So, this measure, as defined in this work, is not intended to improve model training directly for multi-task learning or, at least, not as the primary goal. Furthermore, the aim is to compute this similarity through explainable techniques. 

\section{A multitask similarity measure}
\label{sec:similarity}

\subsection{Notation and terminology}
\label{subsec:notation}

We consider a set of tasks $\mathcal{T}=\{\mathcal{T}_1, \ldots, \mathcal{T}_T\}$. We assume that each task $\mathcal{T}_t\in\mathcal{T}$ has already been trained, resulting in a learned function $f^t(x)\approx E[Y^t|X^t=x]$ where $\mathbf{X}^t=(X_1^t,X_2^t, \ldots, X_{d^t}^t)$ is the vector of $d^t$ predictors for task $\mathcal{T}_t$ and $Y^t\in\mathcal{Y}\subseteq\mathbb{R}$ is a response variable. Each task has an associated dataset $\mathcal{D}_t$ consisting of $n_t\in\mathbb{N}$ samples such that $\mathcal{D}_t = \{\mathbf{x}^t_j=(x_{j,1}^t,\ldots,x_{j,d^t}^t), y_j^t\}_{j=1}^{n_t}$. It is worth noting that this dataset $\mathcal{D}_t$ may differ from the training dataset used to derive the function $f^t(x)$, and it is used to build the ALE curves and compute the weights of the weighted Fr\'echet distance.

$\mathcal{FI}^{t}: \{1,\ldots,d\}\rightarrow [0,1]$ refers to the importance of each variable $X_j$ in the task $\mathcal{T}_t$. This value can be calculated in various ways. There are many techniques to calculate feature importance. Some models allow for direct extraction of the importance of predictor variables, such as linear regression or the Random Forest model \cite{breiman_random_2001}. There are also model-agnostic techniques, such as Permutation Feature Importance (PFI) \cite{fisher_all_2019} or SHAP values \cite{lundberg_unified_2017}. In either case, it should satisfy that $\sum_{j=1}^d\mathcal{FI}^{t}(j) = 1$ for all $\mathcal{T}_t\in\mathcal{T}$. The construction of this function can be derived from any of the available explainable techniques, or it can be created manually. In the latter case, an expert can specify the values to highlight certain features that may be actionable or relevant for the research. Regardless of the approach, the importance values assigned to each variable will significantly influence the measure.

In the simplest case, all features $j$ of each task, i.e.,  $X^{1}_j, \ldots, X^{T}_j$ can express the same characteristic, e.g., arterial blood pressure.  However, while this simplification aids in computation, it is not a strict requirement, meaning that, for example, variable $X_j^{t}$ can express the same concept as variable $X_{j'}^{t'}$ with $j\neq j'$, and the measure defined later is ready to accommodate this issue.

For each variable $j$ of each task $\mathcal{T}_t$, we define a partition $\mathcal{P}_j^t := \{\mathcal{P}_j^t(k) = (z_{k-1,j}^t,z_{k,j}^t]: k = 0, 1\ldots,K^t\}$ consisting of $K^t\in \mathbb{N}$ intervals, where $z_{0,j}^t = min(x_j^t)$ and $z_{K^t,j}^t = max(x_j^t)$. For any $x \in \mathcal{P}_j^t(k)$, the index of the interval into which $x$ falls is defined by $k_j^t(x) := \{k: x \in \mathcal{P}_j^t(k)\}$ and the number of observations that fall into interval $\mathcal{P}_j^t(k)$ is defined by $n_j^t(k) := |\{\{x^t_{ij}\}_{i=1}^{n_t}:x^t_{ij}\in \mathcal{P}_j^t(k)\}|$. For the purpose of weighting the distance function and because tasks can have a different number of observations, it is convenient to compute the proportion of observations that fall into interval $\mathcal{P}_j^t(k)$ as $p_j^t(k):=\frac{n_j^t(k)}{\sum_{k=0}^{K^t}n_j^t(k)}$.

Finally, the subset of $d-1$ predictors of task $\mathcal{T}_t$ excluding $\mathbf{x}^t_{j}$ is defined as $\mathbf{x}_{\backslash j}^t := (x_k^t:k\in \{1,\ldots,d\}\backslash \{j\})$.

\subsection{Accumulated Local Effects (ALE)}
\label{subsec:ale}

Our definition of task similarity is based on the main Accumulated Local Effects (ALE)~\cite{apley_visualizing_2020}. The main idea behind ALEs is to compute a curve that accumulates the averaged local effects of each variable, considering the values of the other variables. 

Although ALEs of second and beyond orders can be computed, in this work, we only consider the main accumulated local effects, i.e., we will compute the ALE curves for each feature separately. The uncentred ALE curve of the variable $j$ and task $\mathcal{T}_t$ is a function 
\begin{equation*}
    \hat{g}_{j,ALE}^t: \mathcal{P}^t_j \rightarrow \mathbb{R} 
\end{equation*}

\noindent such as

\begin{equation*}
    \hat{g}_{j, ALE}^t((z_{k-1,j}^t,z_{k,j}^t]) = \sum_{p=1}^{k}\frac{1}{n^t_j(p)} h^t(j, p)
\end{equation*}

\noindent where

\begin{equation*}
    h^t(j, p) = \sum_{\{i:x_{i,j}^t\in \mathcal{P}_j^t(p)\}}\{f^t(z_{p,j},\mathbf{x}_{i,\setminus j})-f^t(z_{k-1,j},\mathbf{x}_{i,\setminus j})\}.
\end{equation*}

The (centre) ALE main effect so that the ALE function has a mean of 0 with respect to the marginal distribution is calculated by 

\begin{equation}
\label{eq:ALE}
    \hat{f}_{j,ALE}^t(\mathcal{P}_j^t(k)) = \hat{g}_{j, ALE}^t(\mathcal{P}_j^t(k)) - \frac{1}{n_t}\sum_{k=1}^{K^t} n_j^t(k)\hat{g}_{j, ALE}^t(\mathcal{P}_j^t(k))
\end{equation}

In the original paper of ALE plots~\cite{apley_visualizing_2020}, all the examples used the partition $\mathcal{P}_j^t$ based on the quantiles of the empirical distribution. However, in the similarity measure defined here, we prefer to select an equally spaced partition, similar to the standard method to compute a histogram. This choice allows us to capture the shape of the data distribution and determine how to weight the similarity measure.

\subsection{Weighted Fr\'echet Distance}
\label{subsec:wFrechet}

\begin{figure}[!t]
\centering
\includegraphics[width = 0.7\textwidth]{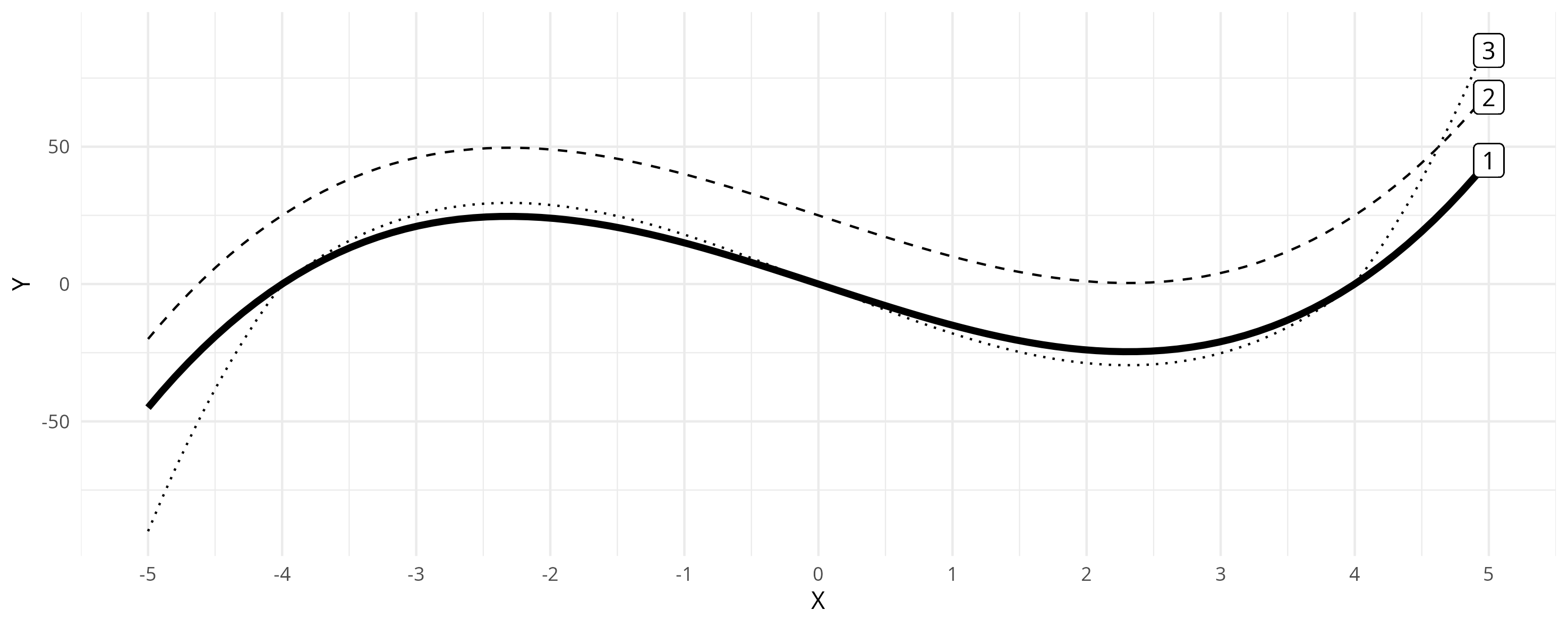}
\caption{ALE curves of different features. Intuitively, curves 1 and 3 are more similar than curves 1 and 2.}
\label{fig:sum-min-frechet}
\end{figure}

The Fr\'echet distance quantifies the similarity between curves by considering both the positioning and sequence of points along the curves. In our case, the curves are the ALE curves that capture the behavior of the model in each variable. 

The estimation of the main ALE curve explained in Section~\ref{subsec:ale} is a polygonal curve so, instead of using the definition of the usual Fr\'echet distance, we employ the discrete version.

In this context, we suppose that (polygonal) ALE curve $\hat{f}_{j,ALE}^{t}$ is defined as $\sigma(\hat{f}_{j,ALE}^{t}) = (u_1,\ldots, u_{K^t})$ such that $u_k = (z_{k,j}^t, \hat{f}_{j,ALE}^{t}((z_{k-1,j}^t,z_{k,j}^t]))$. 

For the Fr\'echet distance, we need to define a coupling between two polygonal curves $\hat{f}_{j,ALE}^{t}$ and $\hat{f}_{j',ALE}^{t'}$ with $t\neq t'$ as a sequence $(u_{a_1}, v_{b1}), (u_{a_2}, v_{b2}),\ldots,(u_{m}, v_{m})$ of distinct pairs from $\sigma(\hat{f}_{j,ALE}^{t})\times\sigma(\hat{f}_{j',ALE}^{t'})$ such that $a_1=1, b_1=1, a_m=K^t, b_m=K^{t'}$ and, for all $i=1,\ldots, q$ we have $a_{i+1}=a_i$ or $a_{i+1}=a_i + 1$, and $b_{i+1}=b_i$ or $b_{i+1}=b_i$

The weighted Fr\'echet Distance is a slight modification of the Discrete Fr\'echet distance~\cite{eiter_computing_1994}. In our notation, the weighted distance between two ALEs curves of two different tasks is as follows:

\begin{equation}
\label{eq:w-frechet-dist}
\delta_F(\hat{g}_{j}^{t},\hat{g}_{j'}^{t'}) := min\{||L||: L \text{ is a coupling between }\hat{g}_{j}^{t}\text{ and } \hat{g}_{j'}^{t'}\}
\end{equation}

\noindent where  $||L|| = \sum_{i=1}^m w(p_{u_{a_i}}^t, p_{v_{b_i}}^{t'})d(u_{a_i},v_{b_i})$, and  coupling $L$ is a sequence $(u_{a_1}, v_{b1}), (u_{a_2}, v_{b2})$.

The function $d$ represents a distance function, most commonly the Euclidean distance.

The weighted Fr\'echet distance can be computed in $\mathcal{O}(pq)$ time~\cite{eiter_computing_1994}. However, if the number of segments is high or varies across tasks, approximation techniques such as Dynamic Time Warping (DTW)\cite{yadav_dynamic_2018} or segment aggregation can be used.

This weighted distance can be seen as a similarity between two variables from different tasks. As mentioned in Section~\ref{subsec:notation}, it is not necessary that the same features occupy the same position in two tasks. In fact, this measure can be applied to two (a priori) unrelated features to identify which feature impacts the response variable similarly. However, if all features are named identically across all tasks (and correspond to the same concept), it simplifies the computation, as we only need to calculate the similarity $d\times (T-1)$ times, as opposed to $d^t$ times (assuming that all tasks have $d$ features). 

The typical formulation of the Fr\'echet distance employs the minimum in Equation~\ref{eq:w-frechet-dist}, although other alternatives exist that utilizes the sum, as in this work. The rationale for this choice is intuitively explained in Figure~\ref{fig:sum-min-frechet}. In this figure, curves labeled 1 and 3 are virtually indistinguishable across most of the interval, with any minor differences attributable to the inherent uncertainty of the models. These curves only diverge at the extremes of the domain so they have to be considered as similar. On the other hand, curves 1 and 2, despite exhibiting a similar pattern, are less similar than curves 1 and 3. If we calculate the Fr\'echet distance between these curves (ignoring the weighted term) using the minimum in Equation~\ref{eq:w-frechet-dist}, we obtain a value of 45 between curves 1 and 2 as well as between curves 1 and 3, despite the differences exposed previously. However, if we use the sum instead of the minimum, the curves 1 and 2 have a Fr\'echet distance of 4229.13 and curves 1 and 3 have a distance of 338.13 highlighting the relevance of using the sum in this measure.

To eliminate the artefactual increase in similarity that occurs when ALE curves are discretised at different resolutions, it is necessary to first project every curve onto a common, task-independent grid. Specifically, for each feature, all observations across tasks are pulled to compute the $K$ empirical quantiles $\{q_1,\dots,q_K\}$, and each task's ALE values are linearly interpolated onto these knots. The result is a set of curves with identical abscissae and bin weights, so the weighted Fr\'echet distance reflects only genuine functional differences rather than arbitrary segment counts. In practice, this resampling step is $O(K)$ per curve and introduces negligible overhead, while ensuring that the multi-task similarity measure remains invariant to uniform curve refinement and numerically stable across datasets.

In the weighted Fr\'echet distance function, we also introduce a weighted function $w: (0,1]\times (0,1]\rightarrow [1, \infty)$. The main aim of this function is to serve as a quantifier of the reliability of each segment of the ALE curve because certain segments may have been calculated with very few observations, leading to potentially unimportant or less accurate estimations of the variable of interest. The exact definition of this weighted function can vary, and it has to be related with the context of the environment of the tasks. As a general function, we propose Equation~\ref{eq:weight}.

\begin{equation} \label{eq:weight}
w(p_{u_{a_i}^t}, p_{v_{b_i}}^{t'}) := \frac{max\{p_{u_{a_i}}^t, p_{v_{b_i}}^{t'}\}}{min\{p_{u_{a_i}}^t, p_{v_{b_i}}^{t'}\}}
\end{equation}

The purpose of the weighted function is to ponder segments of the two ALE curves that we want to compare, ensuring they have sufficient support and are relevant for the comparison. Upon reviewing Figure \ref{fig:sum-min-frechet} again, it becomes apparent that the differences between curves 1 and 3 at the extremes may be attributed to the uncertainty of the models due to the limited support of observations. Therefore, this region should be assigned less weight in the computation of the similarity measure.
Naturally, the function $w$ is symmetrical, i.e., $w(p_{u_{a_i}^t}, p_{v_{b_i}}^{t'})=w(p_{v_{b_i}}^{t'}, p_{u_{a_i}^t})$

Algorithm~\ref{alg:weight-dist} shows the Weighted Fr\'echet distance in pseudo-code.

\begin{algorithm}
\caption{Weighted Discrete Fr\'echet Distance}
\begin{algorithmic}[1]
\REQUIRE Two polygonal ALE curves 
\(\sigma_1 = [(x_1,u_1),\dots,(x_p,u_p)]\) with weights \(w_1[1..p]\),\\
\(\sigma_2 = [(y_1,v_1),\dots,(y_q,v_q)]\) with weights \(w_2[1..q]\).
\ENSURE \(\delta_F(\sigma_1,\sigma_2)\).

\STATE Initialize \(D[1..p][1..q] \leftarrow +\infty\).
\FOR{\(i = 1\) to \(p\)}
  \FOR{\(j = 1\) to \(q\)}
    \STATE \( \omega \leftarrow \dfrac{\max(w_1[i],\,w_2[j])}{\min(w_1[i],\,w_2[j])}\)
    \STATE \( d \leftarrow \sqrt{(x_i - y_j)^2 + (u_i - v_j)^2}\)
    \IF{\(i=1\) \textbf{and} \(j=1\)}
      \STATE \(D[i][j] \leftarrow \omega \cdot d\)
    \ELSE
      \STATE \(\mathrm{best} \leftarrow \min\bigl(D[i-1][j],\,D[i][j-1],\,D[i-1][j-1]\bigr)\)
      \STATE \(D[i][j] \leftarrow \mathrm{best} + \omega \cdot d\)
    \ENDIF
  \ENDFOR
\ENDFOR
\RETURN \(D[p][q]\)
\end{algorithmic}
\label{alg:weight-dist}
\end{algorithm}

\subsection{Multi-task similarity measure}
\label{subsec:similarity}

The weighted Fr\'echet distance enables us to compare two features from two different tasks. This distance can serve as the foundation for defining a measure that allows us to assess the similarity between two tasks, but before we can formulate the definition of this measure, there are a couple of aspects that we need to consider. On the one hand, one of the primary objectives of this measure is to uncover how specific features affect the target similarly. In this context, it is not necessary to compare all the features present in the training dataset for both tasks. Furthermore, there might be features available in one task that are not present in the other task. To address this, the measure can consider only some of the features. Another reason could be that we want to measure the similarity between tasks only for actionable features. On the other hand, generally, we do not assume that the $j$-th feature in both tasks expresses the same characteristic, i.e., the first feature in one task might represent the blood pressure of the patient, whereas this feature could be the fourth one in another task.

The multi-task similarity measure between a task of interest $\mathcal{T}_t$ and other task $\mathcal{T}_{t'}$ is a function 
\begin{equation*}
    \delta_t: \mathcal{T}\setminus\{\mathcal{T}_t\} \rightarrow \mathbb{R}^+    
\end{equation*}

\noindent is defined as

\begin{equation}\label{eq:sim-measure}
\delta_t(t') := \sum_{j=1}^{d^t} \mathcal{FI}^t(j)\underset{j'\in \{1, \ldots, d^{t'}\}}{\min} \delta_F(\hat{f}_{j,ALE}^{t},\hat{f}_{j',ALE}^{t'})
\end{equation}

The pseudocode for the Multi-task Similarity measure is shown in Algorithm~\ref{alg:multitask-sim}.

\begin{algorithm}
\caption{Multi-Task Similarity Measure}
\begin{algorithmic}[1]
\REQUIRE Tasks \(t_0, t_1\); feature sets \(\mathcal{F}_{t_0}, \mathcal{F}_{t_1}\)\\
ALE curves \(\mathrm{ALE}[t][f]\), weights \(\mathrm{prop}[t][f]\), importances \(\mathrm{imp}[t][f]\).
\ENSURE \(\displaystyle \delta(t_0,t_1)\).

\STATE \(\mathrm{sim} \leftarrow 0\)
\FOR{each feature \(f\in \mathcal{F}_{t_0}\)}
  \STATE \(\mathrm{minDist} \leftarrow +\infty\)
  \FOR{each feature \(g\in \mathcal{F}_{t_1}\)}
    \STATE \(d \leftarrow \text{WeightedFrechetDistance}\bigl(\mathrm{ALE}[t_0][f],\,\mathrm{prop}[t_0][f],\,
                                                 \mathrm{ALE}[t_1][g],\,\mathrm{prop}[t_1][g]\bigr)\)
    \STATE \(\mathrm{minDist} \leftarrow \min(\mathrm{minDist},\,d)\)
  \ENDFOR
  \STATE \(\mathrm{sim} \leftarrow \mathrm{sim} + \mathrm{imp}[t_0][f]\times \mathrm{minDist}\)
\ENDFOR
\RETURN \(\mathrm{sim}\)
\end{algorithmic}
\label{alg:multitask-sim}
\end{algorithm}

Note that although the weighted Fr\'echet distance is symmetrical, being Equation~\ref{eq:weight} symmetrical, the multi-task similarity measure might not be symmetrical. This is due to the influence of the importance weights in task $t$, which may not coincide with the feature importance weights of task $t'$.

\subsubsection{Multi-task similarity measure including model performance}
\label{subsubsec:similarity-performance}

The similarity defined in Equation~\ref{eq:sim-measure} can be misleading when tasks have very different predictive quality.  
Such disparities arise when (i) the model assigned to a task lacks sufficient capacity—e.g.\ using a linear learner for data generated by a nonlinear process—or (ii) some tasks are intrinsically harder.

Throughout the paper we interpret the multi-task similarity measure $\delta_t(t')$ as $\delta_t(t') = 0$ means identical ALE profiles, while larger values indicate tasks that diverge more strongly.  
This convention matters because the corrective factor introduced below \emph{shrinks} large values when they are attributable to unequal model performance.

Poor performance should not automatically translate into perceived task dissimilarity.  
To account for this, we multiply the raw discrepancy by a weight $\gamma_t(t')\!\in[0,1]$ that depends on the empirical losses $L(t)$ and $L(t')$ (smaller is better):

\begin{equation}
  \gamma_t(t') \;=\; f\!\bigl(L(t),L(t')\bigr),
\end{equation}

\noindent with the desideratum that $\gamma_t(t')\approx 1$ when losses are comparable and $\,\gamma_t(t')\!\downarrow 0\,$ as their ratio grows.  
The \emph{performance-weighted} similarity is then
\begin{equation}\label{eq:sim-performance}
    \delta_t^{\ast}(t') \;:=\; \gamma_t(t') \cdot \delta_t(t') .
\end{equation}

A simple and scale-free choice that meets the desiderata is
\begin{equation}
  \gamma_t(t') 
  \;=\; 
  \frac{\min\{L(t),\,L(t')\}}
       {\max\{L(t),\,L(t')\}+\varepsilon},
  \qquad
  \varepsilon>0,
\end{equation}
where $\varepsilon\!=\!10^{-8}$ (unless stated otherwise) guarantees numerical stability when both tasks fit almost perfectly.

If \emph{both} losses are large, the ratio above is close to~1 even though neither model is trustworthy.  
We therefore flag any task with $L(t)>\tau$ (e.g., $\tau$ could be the median loss across tasks) and recommend inspecting or retraining the corresponding model before drawing conclusions from $\delta_t^{\ast}$.

\subsubsection{Limitations and recommendations}
\label{subsubsec:limitations}

The Multi-task similarity measure developed in this work has certain limitations. We highlight these limitations here and provide recommendations to help mitigate these issues.

The ALE curves used for the multi-task similarity computations rely on first-order ALE (i.e., they only consider each variable individually). If the true signal is an interaction (e.g., $X_1\times X_2$), both one-way ALEs can be flat even though the joint surface differs sharply between tasks. The Fr\'echet distance can be only used between curves parametrized in one dimension (as first-order ALE curves), so the measure cannot be applied to second or higher order ALE curves. In this case, the recommendation is to first learn a shallow autoencoder $Z = f_\theta(X)$ on the union of tasks and compute the ALE curves on $Z$ instead of raw $X$. The encoder allows for the alignment of heterogeneous feature sets and preserves interactions captured in $Z$.

However, there is a word of caution regarding this procedure. If the intended purpose is to assess the similarities between tasks in an explainable manner, the output of the encoder may not preserve the interpretability. Nevertheless, the measure values can still be used as a similarity metric between tasks and can serve, for instance, in clustering purposes or as a basis for multi-task learning paradigms.

To address this limitation while preserving interpretability, future work will explore extending the similarity measure to integrate selective second-order effects. One avenue is to compute second-order ALE curves for pre-identified key interaction pairs (via domain knowledge or model-based interaction importance) and aggregate similarity scores from both first-order and selective second-order ALEs.

Another limitation that is important to note is the computational complexity of the measure. There are three main sources contributing to this complexity: (i) the computation of ALE curves, (ii) the calculation of the weighted Fr\'echet distance between ALE curves, and (iii) the aggregation of similarity scores across all tasks and features. In the worst case, the current implementation scales as $\mathcal{O}(T^2dK^2)$ where $T$ is the number of tasks, $d$ the number of features, and $K$ the number of ALE segments.

The first source is related to the size of the data and the speed at which the trained model can produce predictions. A simple optimization is to compute the ALE curves using only a representative sample of the data, which can significantly reduce computational load without severely affecting accuracy.

The second source stems from the Fr\'echet distance computation between ALE curves, which scales with the number of segments in the curves and the number of pairwise computations. To alleviate this, one could (a) use curves with a reduced or adaptive number of segments, (b) apply approximation techniques or pruning heuristics for the Fr\'echet distance, or (c) parallelize the pairwise distance calculations.

The third source arises from computing the similarity between each task pair across features. As these computations are independent across task pairs, the most recommended solution is to perform the calculations using parallel processing to take advantage of modern multi-core or distributed computing environments. For a very large number of tasks $T$, we recommend first clustering tasks with a lightweight metric  (e.g., cosine similarity of feature-importance vectors) and computing the full measure only within or between selected clusters, reducing the effective number of pairwise workload from $\mathcal{O}(T^2)$  to roughly $\mathcal{O}(T^2_c)$, where $T_c \ll T$ is the average cluster size.

Although the similarity measure has so far been presented for tabular inputs, many real-world tasks involve non-tabular data, such as image, text, or audio signals. A natural remedy is to use an encoder to transform such data into a latent representation and compute similarity based on these latent features, as proposed at the beginning of this section for heterogeneous task structures. However, this approach introduces a significant limitation: it compromises the interpretability of the similarity measure, as latent features typically lack direct semantic meaning.

To overcome this challenge while preserving compatibility with the ALE Fr\'echet distance framework, we outline three interpretable transformation strategies that enable the application of our similarity measure to non-tabular data:

\begin{enumerate}
\item \textbf{Concept-bottleneck encoders.}
We propose employing encoders that explicitly predict human-interpretable concepts in their penultimate layer (e.g., edge density, tumor shape, or sentiment polarity) prior to producing the final output as Concept Bottlenecks Models \cite{koh_concept_2020}. These concept activations provide a semantically labeled tabular representation on which first-order ALE curves can be computed, thereby retaining the transparency and explainability of the similarity analysis.

\item \textbf{Domain-specific feature libraries.}
In many fields, there exist well-established libraries of handcrafted, physically meaningful descriptors—such as radiomic features for medical images~\cite{lambin_radiomics_2017}, psycholinguistic or sentiment markers for text~\cite{tausczik_psychological_2010}, and spectral coefficients for audio signals~\cite{davis_comparison_1980}. By extracting these features prior to model training, we avoid reliance on black-box embeddings and ensure that the similarity measure is grounded in features with clear interpretability.

\item \textbf{Attention-pooled region features.}
Transformer-based architectures naturally provide attention maps that highlight regions (e.g., image patches or text tokens) most influential to a prediction~\cite{chefer_transformer_2021}. We propose aggregating these attentions into $K$ coherent super-pixels (in vision) or phrase clusters (in NLP), resulting in $K$ interpretable region scores. ALE curves computed on these scores illustrate how specific regions modulate outputs, while weighted Fr\'echet distances on these curves quantify inter-task differences. The original attention maps can further serve as visual explanations that complement the numerical similarity measure.

\end{enumerate}
Each of these strategies offers a pathway to extend our method to non-tabular data without sacrificing interpretability. Nevertheless, we acknowledge that the transformation step introduces its own assumptions and potential biases, which should be carefully considered in practical applications. We highlight these directions as promising areas for future research, particularly for integrating explainability in complex, multimodal task similarity analysis.

Finally, we acknowledge an inherent limitation in the proposed interpretability framework. The multitask similarity measure is built on Accumulated Local Effects (ALE) curves, which depict how individual features influence model predictions. When the true feature‒output relationship is highly non-linear, these curves often exhibit high-frequency oscillations that can overwhelm domain experts and obscure the underlying pattern. We address this concern in two complementary ways. (i) Encoder-based filtering as described earlier in this section, an encoder network can learn a latent representation that captures the salient, shared structure of the ALE curves while discarding task-specific noise. (ii) Spline smoothing of ALE curves. Alternatively, each ALE curve can be regularized directly: we fit a surrogate spline $\hat{g}(x)$ that minimizes a roughness-penalized $L_{2}$ loss, thereby preserving the global trend but suppressing idiosyncratic wiggles. Both strategies attenuate fine-grained fluctuations that do not contribute meaningfully to the multitask similarity, yielding explanations that remain faithful yet are considerably easier for experts to interpret.

\section{Empirical Work}
\label{sec:empiricalWork}

In this section, we apply the similarity measure defined in Section~\ref{subsec:similarity} in three different contexts. 

Firstly, we simulate toy data from various tasks, each with a few features, to comprehend the behavior of the similarity measure. All tasks share features arranged in the same order, making them directly comparable.

Secondly, we apply the multi-task similarity measure to a real data set containing information about Parkinsons' patients. This demonstrate the measure's applicability in a practical context and allow us to discuss its utility in understanding task relationships.

Lastly, we use the multi-task similarity in a bike-sharing dataset from the public service BiciMad in Madrid, Spain. In this case, each task corresponds to one of the 264 stations available to users, with the goal of predicting the number of bikes used per hour based on temporal features (e.g., month, day of the week, hour) and weather conditions (e.g., humidity, temperature). For this dataset, we employ a hybrid parameter-sharing multi-task deep learning algorithm.

In the first dataset, due to its simplicity, we aimed to test whether the tasks intuitively considered similar align with the similarity detected by the measure. In contrast, the other datasets present more complex scenarios, allowing us to demonstrate the full potential of the similarity measure. These cases highlight the benefits and utility of the measure in detecting both general and task-specific behaviors across the datasets.

The details of the model setups can be found respectively in \ref{appendix:dataset-1}, \ref{appendix:parkinson}, \ref{appendix:bikes} and \ref{appendix:celeba}. The implementation of the measure is available at github.com/papabloblo/multi-task-similarity.

\subsection{Synthetic Dataset}
\label{subsec:syntheticdata1}

The synthetic dataset simulated consists of five tasks, each with five features and $10,000$ observations. For the multitask-measure developed in this work, three key factors must be considered: (1) the distribution of each predictive feature, (2) the relative importance of each feature in the predictive model, and (3) the ALE curves. 

\begin{figure*}[!t]
\centering
\includegraphics[width = 0.9\textwidth]{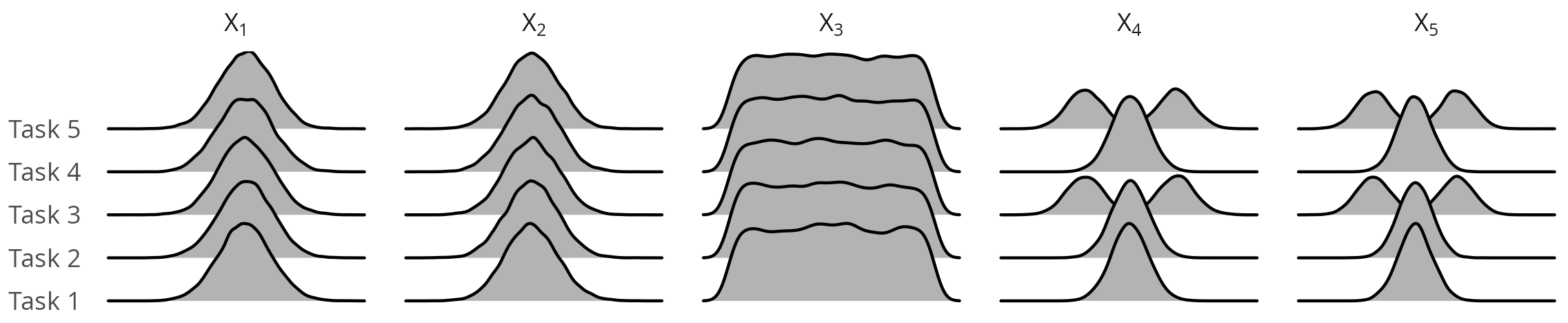}
\caption{Density of each predictor variable for each task in Synthetic Dataset 1.}
\label{fig:synth1-density}
\end{figure*}

Figure~\ref{fig:synth1-density} shows the density of each variable across all tasks. It is evident that the tasks exhibit very similar densities for variables $X_1$, $X_2$ and $X_3$, while differences are observed in variables $X_4$ and $X_5$. This indicates that even if the patterns of two ALE curves are only similar within a narrow segment, a segment containing the majority of the data points will significantly contribute to a higher similarity. In the definition of the measure, this corresponds to a lower value. The variables $X_1$ and $X_2$ are the only predictive variables that are related, following a bivariate normal distribution, whereas the remaining predictors are unrelated to each other. Details of the data simulation are provided in \ref{appendix:dataset-1}.

Each one of the five tasks was trained independently in a single-task learning fashion using a Random Forest model~\cite{breiman_random_2001} and the importance measure of each feature was calculated according to the method described by Breiman~\cite{breiman_random_2001}, with the results transformed so that the sum of the feature importance for each of the tasks is $1$.

\begin{figure*}[!t]
\includegraphics[width = \textwidth]{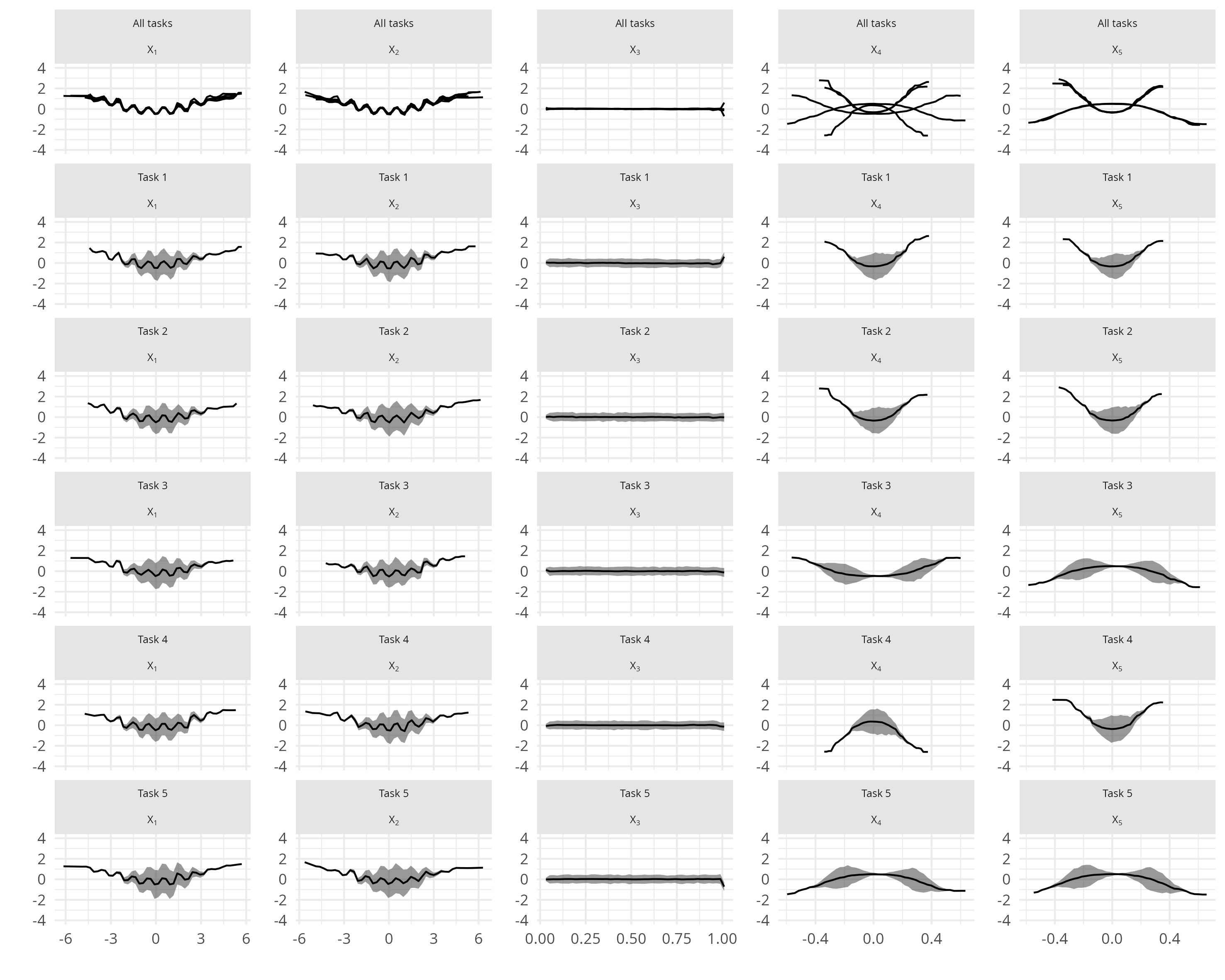}
\caption{ALE Curves for the Synthetic Dataset 1. The first row shows all the ALE curves of the different tasks for the same feature. The rest of the rows show ALE curves for each task and feature.}
\label{fig:synth1-ale}
\end{figure*}

Afterwards, the ALE curve of each feature and task was calculated using an equally spaced partition comprising 50 intervals. Figure \ref{fig:synth1-ale} shows the ALE plots for each variable and task, demonstrating similarities in behavior among features across tasks. The marginal distribution of each variable is depicted as the area below the ALE curves. This has a huge impact on the similarity value. 

Variables $X_1$, $X_2$, and $X_3$ exhibit nearly identical curves across tasks, albeit for different reasons. In contrast, the variable $X_3$ exhibits a flat curve across tasks indicating, as previously observed, that this predictor variable is unrelated to the outcome. Finally, features $X_4$ and $X_5$ exhibit different behaviors across tasks as illustrated by the ALE plots in Figure~\ref{fig:synth1-ale}.

With these plots, and considering that variables $X_1$, $X_2$, and $X_3$ exhibit similarity across tasks and, therefore, do not contribute to making tasks more or less similar, we can visually determine which tasks are similar. Based on the shape of the curves, it is obvious that tasks 1 and 2 are very similar with respect to all the variables. However, the remaining tasks are only partially similar to each other due to the mixture of shapes in the ALE plots of variables $X_4$ and $X_5$, along with the underlying distribution of the predictor variables. This emphasizes the importance of having an objective measure of the similarity between tasks that can consider all the details.

\begin{table}
\small
\caption{Multi-task similarity measure values tasks in Synthetic Dataset 1. Additionally, it presents the weighted Fr\'echet distance between each variable and task, with the last column indicating the feature importance in each task.}
\centering
\begin{tabular}[t]{rccccc|c}
\toprule
 & Task 1 & Task 2 & Task 3 & Task 4 & Task 5 & Imp.\\
\midrule
\multicolumn{7}{l}{\textbf{Task 1}}\\
\hspace{1em}$X_1$ & - & \textbf{10.41} & 14.31 & 13.03 & 11.71 & 0.19\\
\hspace{1em}$X_2$ & - & \textbf{9.46} & 15.53 & 13.17 & 14 & 0.19\\
\hspace{1em}$X_3$ & - & \textbf{1.75} & 2.22 & 2.42 & 3.19 & 0.09\\
\hspace{1em}$X_4$ & - & \textbf{7.89} & 53.25 & 133.27 & 107.53 & 0.27\\
\hspace{1em}$X_5$ & - & 4.73 & 123.69 & \textbf{3.44} & 117.92 & 0.26\\
\midrule
\hspace{1em}Similarity & - & \textbf{7.33} & 52.35 & 42.22 & 64.84 & \\
\midrule
\multicolumn{7}{l}{\textbf{Task 2}}\\
\hspace{1em}$X_1$ & \textbf{10.41} & - & 13.28 & 11.19 & 18.33 & 0.19\\
\hspace{1em}$X_2$ & \textbf{9.46} & - & 20.99 & 11.32 & 10.68 & 0.19\\
\hspace{1em}$X_3$ & 1.75 & - & 1.37 & \textbf{1.22} & 1.89 & 0.09\\
\hspace{1em}$X_4$ & \textbf{7.89} & - & 50.5 & 134.59 & 113.98 & 0.27\\
\hspace{1em}$X_5$ & \textbf{4.73} & - & 129.62 & 5.08 & 140.81 & 0.27\\
\midrule
\hspace{1em}Similarity & \textbf{7.29} & - & 54.68 & 41.78 & 73.72 & \\
\midrule
\multicolumn{7}{l}{\textbf{Task 3}}\\
\hspace{1em}$X_1$ & 14.31 & \textbf{13.28} & - & 14.5 & 13.73 & 0.19\\
\hspace{1em}$X_2$ & \textbf{15.53} & 20.99 & - & 24.22 & 25.47 & 0.19\\
\hspace{1em}$X_3$ & 2.22 & 1.37 & - & \textbf{0.93} & 1.66 & 0.09\\
\hspace{1em}$X_4$ & 53.25 & \textbf{50.5} & - & 183.03 & 90.73 & 0.27\\
\hspace{1em}$X_5$ & 123.69 & 129.62 & - & 123.33 & \textbf{2.51} & 0.27\\
\midrule
\hspace{1em}Similarity & 53.03 & 54.64 & - & 89.02 & \textbf{32.31} &\\
\midrule
\multicolumn{7}{l}{\textbf{Task 4}}\\
\hspace{1em}$X_1$ & 13.03 & \textbf{11.19} & 14.5 & - & 12.31 & 0.19\\
\hspace{1em}$X_2$ & 13.17 & 11.32 & 24.22 & - & \textbf{9.51} & 0.19\\
\hspace{1em}$X_3$ & 2.42 & 1.22 & \textbf{0.93} & - & 1.56 & 0.08\\
\hspace{1em}$X_4$ & 133.27 & 134.59 & 183.03 & - & \textbf{41.08} & 0.28\\
\hspace{1em}$X_5$ & \textbf{3.44} & 5.08 & 123.33 & - & 126.64 & 0.26\\
\midrule
\hspace{1em}Similarity & \textbf{42.76} & \textbf{42.76} & 90.49 & - & 49.1 & \\
\midrule
\multicolumn{7}{l}{\textbf{Task 5}}\\
\hspace{1em}$X_1$ & \textbf{11.71} & 18.33 & 13.73 & 12.31 & - & 0.19\\
\hspace{1em}$X_2$ & 14 & 10.68 & 25.47 & \textbf{9.51} & - & 0.18\\
\hspace{1em}$X_3$ & 3.19 & 1.89 & 1.66 & \textbf{1.56} & - & 0.09\\
\hspace{1em}$X_4$ & 107.53 & 113.98 & 90.73 & \textbf{41.08} & - & 0.26\\
\hspace{1em}$X_5$ & 117.92 & 140.81 & \textbf{2.51} & 126.64 & - & 0.27\\
\midrule
\hspace{1em}Similarity & 64.94 & 73.34 & \textbf{31.76} & 49.17 & - & \\
\bottomrule
\end{tabular}
\label{table:synth1-similar-tasks-var}
\end{table}

\begin{figure*}[!t]
\centering
\includegraphics[width = .6\textwidth]{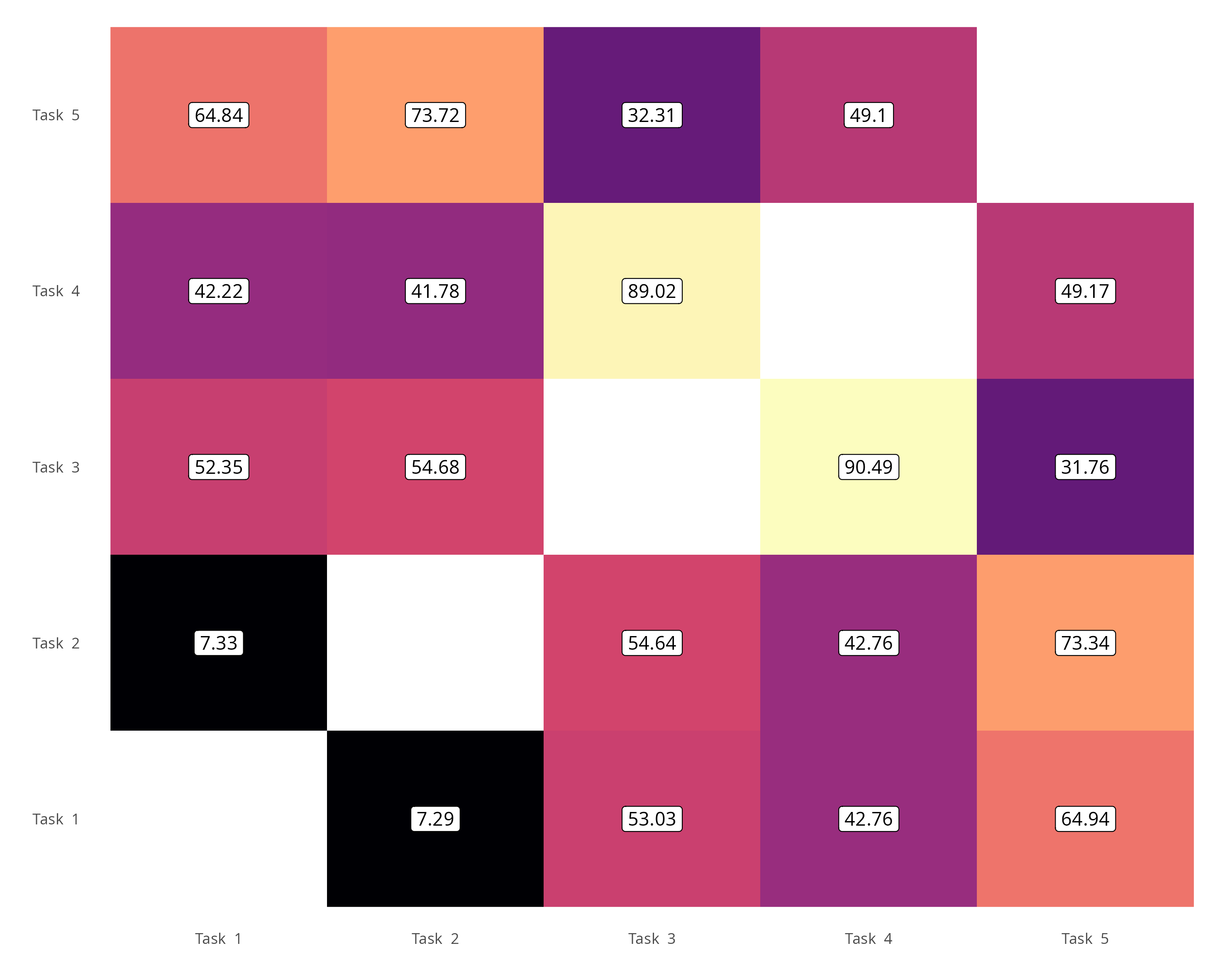}
\caption{Multi-task similarity values.}
\label{fig:synth1-heatmap}
\end{figure*}

Although the multi-task similarity measure described in this paper provides a value indicating the similarity between tasks, to compute this value, we first have had to calculate the weighted Fr\'echet distance between one variable of each task with respect the variables of the other tasks as described in Equation~\ref{eq:w-frechet-dist}. These intermediate computations help us understand the behavior of the measure and are displayed in Table~\ref{table:synth1-similar-tasks-var}, along with the importance of each variable and the final multi-task similarity measure. It should be noted that, for simplicity, we assumed that all features are named identically across all tasks, and we only considered relations between variables named identically. Consequently, this implies that we only need to calculate the similarity between variables $X_1$ across tasks, followed by $X_2$, and so forth. Remember that, by the definition of the measure, lower values indicate that the tasks considered are more similar. A summarize of the similarity between tasks is displayed in Figure~\ref{fig:synth1-heatmap}.

According to Table~\ref{table:synth1-similar-tasks-var}, variables $X_1$, $X_2$, and $X_3$ exhibit analogous values of similarity across tasks with slight variations. This follows the expected behavior because all the tasks follow the same relationship of $X_1$ and $X_2$ to the outcome, and the variable $X_3$ is unrelated to the outcome. Notice that the weighted Fr\'echet distance between variables $X_3$ has lower values than the other variables. This underscores the relevance of considering not only the distance between variables but also the importance of the variables in the multi-task similarity measure. Although the predictor variable $X_3$ seems to have a similar pattern in all tasks, assessing its importance can capture the insignificance of this feature in the prediction (we know that, in reality, there is no relation with the outcome) and weigh its contribution to the multi-task similarity computation. 

The most significant differences are observed between variables $X_4$ and $X_5$. It is evident that the shape of the curves is not the only determinant for the weighted Fr\'echet distance. As anticipated in the visual inspection, both tasks~1 and~2 have the lowest values of the multi-task similarity measure and in the majority of the weighted Fr\'echet distances between variables. Therefore, the results of the measure follow the intuition. 

Task 3 exhibits two curves with opposite patterns in variables $X_4$ and $X_5$. Variable $X_5$ has a very similar pattern compared to the same variable in Task 5, and the weighted Fr\'echet distance supports this, resulting in a much lower value compared to the other tasks. However, the variable $X_4$ of Task 3 has a pattern that could be similar to Task 1 or 2, although it has a different distribution. This is reflected by the values of the weighted distance, as we obtain similarly low values compared to Tasks 1 and 2, but these values are much bigger compared to other patterns that are more similar, such as between Tasks 1 and 2. Ultimately, considering the weighted Fr\'echet distance of each variable and the importance of each feature, the multi-task similarity measure indicates that Task 5 is the most similar task. The remaining results can be explored in the values collected in Table \ref{table:synth1-similar-tasks-var}.

\begin{figure*}[!t]
\includegraphics[width = \textwidth]{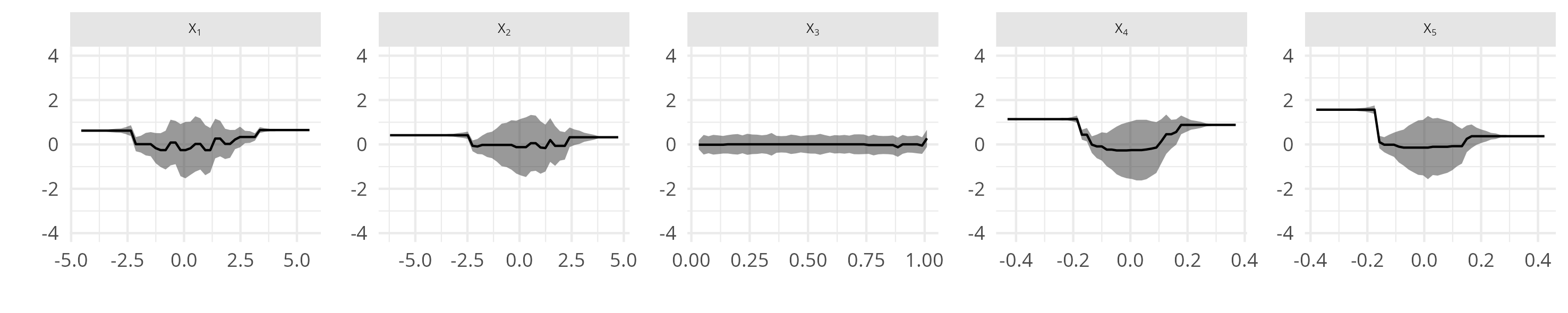}
\caption{ALE Curves for the task 6.}
\label{fig:synth3-ale}
\end{figure*}

As discussed in Section \ref{subsubsec:similarity-performance}, an alternative version of the multi-task similarity measure can be used, incorporating the predictive quality of each task through Equation \ref{eq:sim-performance}. To demonstrate the usefulness of this approach, we simulated new data following the same distribution as Task 1 in the toy dataset, but training a deliberately limited model (specifically, a Random Forest model with only 3 trees and a minimum node size of 750 observations). The ALE curves obtained for this new task (Figure~\ref{fig:synth3-ale}) show that, although there are noticeable differences compared to those of Task 1 (Figure~\ref{fig:synth1-ale}), the overall behavior remains largely similar.

Table~\ref{table:mts-with-performance} presents the multi-task similarity measure between tasks, both with and without incorporating the performance-based scaling factor. The reference task is indicated by the row, and the compared task by the column. Values in parentheses reflect the similarity after adjusting for predictive quality. Notably, the similarities involving Task 6 (the poorly performing task) are considerably reduced when the performance factor is included—demonstrating that the similarity measure appropriately down weights comparisons involving low-quality models. For example, the similarity between Task 1 and Task 6 decreases from 29.73 to 21.34, and between Task 2 and Task 6 from 35.68 to 28.39. Despite the limited model used for Task 6, the similarity values (both raw and scaled) indicate that Task 6 is more similar to Task 1 than to any other task, which is expected given that both tasks share the same data distribution. This further supports the effectiveness of the performance-aware similarity adjustment in preserving meaningful relationships while penalizing unreliable comparisons.

\begin{table}
\centering
\scriptsize
\caption{Multi-task similarity measure between a reference task (rows) and another task (columns). The values in parentheses refer to the measure incorporating the performance-based scaling factor.}
\begin{tabular}[t]{r|cccccr|cccccr|cccccr|cccccr|cccccr|ccccc}
\toprule
 & Task 1 & Task 2 & Task 3 & Task 4 & Task 5 & Task 6\\
\midrule
Task 1 & - & \textbf{7.33} (6.61) & 52.35 (40.91) & 42.22 (34.9) & 64.84 (43.88) & 29.73 (21.34)\\
Task 2 & \textbf{7.29} (6.57) & - & 54.68 (47.37) & 41.78 (38.28) & 73.72 (55.31) & 35.68 (28.39)\\
Task 3 & 53.03 (41.44) & 54.64 (47.33) & - & 89.02 (84.16) & \textbf{32.31} (27.98) & 44.55 (40.93)\\
Task 4 & \textbf{42.76} (35.35) & \textbf{42.76} (39.18) & 90.49 (85.55) & - & 49.1 (40.21) & 55.04 (47.8)\\
Task 5 & 64.94 (43.96) & 73.34 (55.02)& \textbf{31.76} (27.51) & 49.17 (40.26) & - & 57.32 (54.04)\\
Task 6 & \textbf{32.1} (23.04) & 36.92 (29.38) & 48.35 (44.42) & 62.96 (54.68) & 65.73 (61.97) & -\\
\bottomrule
\end{tabular}
\label{table:mts-with-performance}
\end{table}

\subsection{Parkinson dataset}
\label{subsec:realdata}

We also apply the multi-task similarity measure developed to the Parkinson dataset\footnote{\url{https://archive.ics.uci.edu/dataset/189/parkinsons+telemonitoring}}~\cite{athanasios_tsanas_parkinsons_2009}, a real-world dataset containing information about 42 patients. The goal is to predict the disease symptom score of Parkinson at different times using 19 biomedical features. In the multi-task scenario, each of the 42 patients represents a task. The dataset consists of a total of 5,875 records, and each patient (task) has approximately 200 data points. Although it is not the main goal of this work, this dataset is used as a benchmark in several multi-task learning algorithms~\cite{zhang_survey_2022}.

\begin{figure}[ht]
\includegraphics[width = .7\columnwidth]{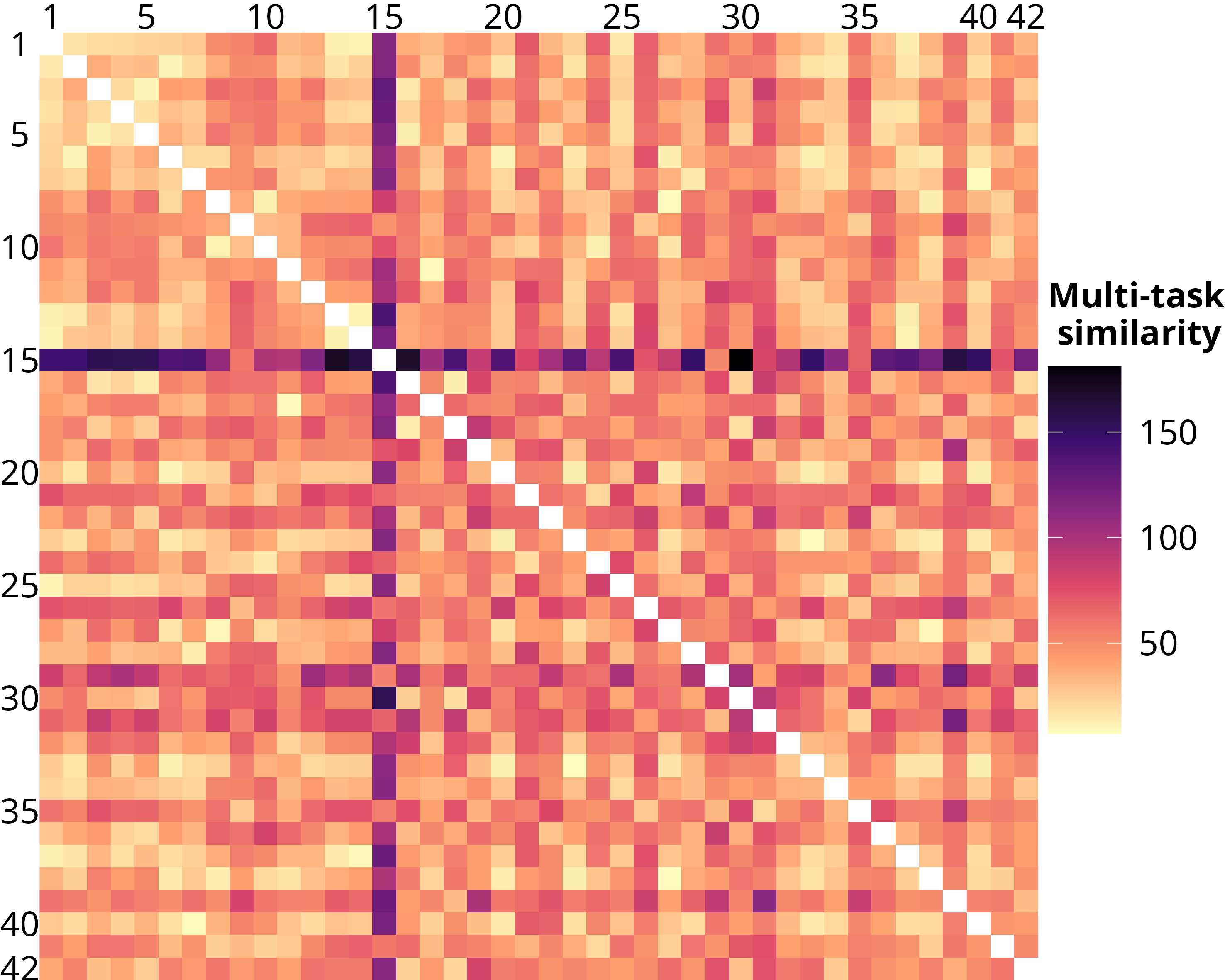}
\centering
\caption{Matrix of multi-task similarity between patients. The matrix must be interpreted as the similarity between a patient in a column with a patient in a row.
}
\label{fig:matrix_sim_real-data}
\end{figure}

Similarly to the approach followed in the previous synthetic dataset (Section~\ref{subsec:syntheticdata1}), we have trained a separate random forest model for each patient (task). Figure~\ref{fig:matrix_sim_real-data} shows the multi-task similarity measure between patients in a matrix format, aiding in the exploration of the results and the extraction of conclusions. As the matrix shows, we can use the multi-task measure to identify patients with different behavior compared to the others, not by using the raw observations, but by employing a model that summarizes the general behavior of the tasks (or patients in this context). For example, patient 14 has lower values of similarity (in general) than patient 15, indicating that patient 15 may have special characteristics that make them unique and different from the other patients in the sample. Patient 14, on the other hand, may exhibit a more homogeneous behavior with the rest of the patients. 

\begin{figure*}
\includegraphics[width = 0.8\textwidth]{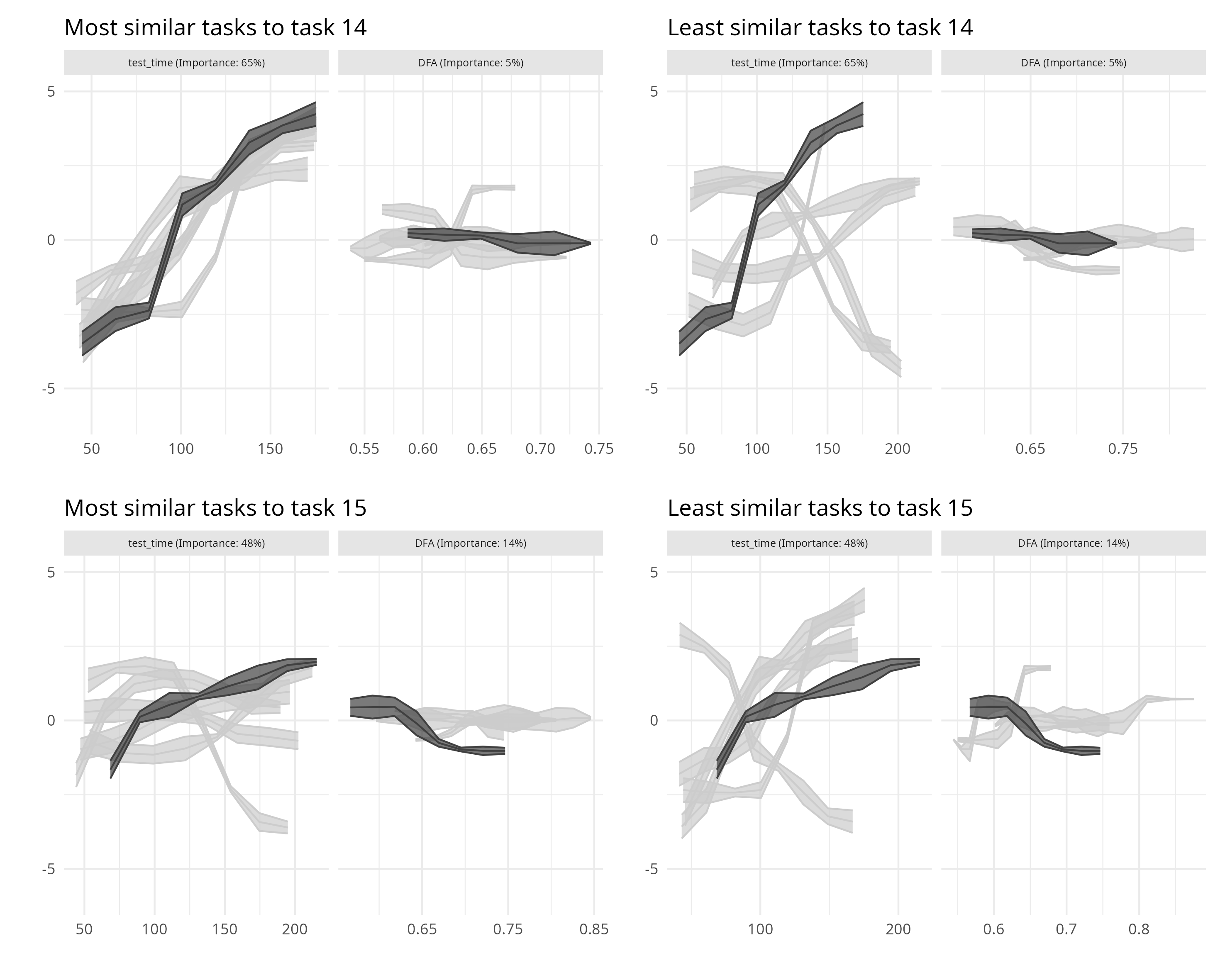}
\centering
\caption{ALE plots for the five most and least similar tasks to Tasks 14 and 15. The bold line represents the task of interest. In the figure, only the two most important variables for each task of interest are shown.}
\label{fig:most-least-sim}
\end{figure*}

Table~\ref{table:real-data} shows the five highest and five lowest similarity values between patients 14 and 15 with respect to the other 41 patients. Figure~\ref{fig:most-least-sim} shows the ALE plots of the two most important variables for each task among the five most and least similar patients (tasks) to patients 14 and 15. As anticipated in the similarity values in Table~\ref{table:real-data}, it can be observed that the tasks similar to Task 14 have a homogeneous increasing pattern in the most important variable \textit{test\_time} (time since recruitment into the trial), in contrast to the most similar tasks to Task 15, where there is a heterogeneous pattern. Furthermore, in the second most important variable (\textit{DFA} or signal fractal scaling exponent), it is clearer that patient 15 exhibits a very different response compared to the other patients considered in the study sample.

With this application, we can see the potential of the measure to serve as an exploratory tool to comprehend not only the general behavior of the tasks but also the relationship and the differences between them.

\begin{table}
\caption{Multi-task similarity (MTS) between the features of Task 1 and the features of the other tasks.}
\centering
\begin{tabular}[t]{c|c|c|c}
\toprule
\multicolumn{2}{c|}{Patient 14} & \multicolumn{2}{c}{Patient 15}\\
\midrule
Patient & MTS & Patient & MTS\\
37 & 9.33 & 9& 47.16\\
13 & 11.52 & 29& 55.64\\
1 & 11.66 & 35& 56.24\\
4 & 20.43 & 41& 58.90\\
25 & 22.21 & 26& 60.87\\
... & ... & ...& ...\\
21 & 76.72 & 4& 126.17\\
31 & 81.54 & 3& 129.56\\
26 & 88.39 & 16& 136.62\\
29 & 98.40 & 13& 139.63\\
15 & 159.31 & 30& 154.93\\
\bottomrule
\end{tabular}
\label{table:real-data}
\end{table}

\subsection{Bike-sharing BiciMad dataset}
\label{subsec:bicimad}

This dataset contains information about the public bike-sharing system BiciMad\footnote{\url{https://www.bicimad.com/en/home}}, which operates in Madrid, Spain. The records span between the years 2021 to 2023 and include data from 264 stations where bikes can be locked. The objective is to predict the number of bikes unlocked at each station within a one-hour period. The dataset comprises slightly more than 2 million rows and 10 features, including temporal variables such as month, weekday, and hour, as well as weather conditions such as precipitation and humidity. In~\ref{appendix:bikes}, Figure~\ref{fig:mapa-estaciones} displays the locations of the stations in Madrid, Spain.

In this case, the model used is a hybrid-parameter sharing deep learning model with three types of layers: (1) layers shared across all tasks with the same parameters (hard parameter sharing), (2) task-specific layers with parameters encouraged to be similar by applying a regularization constraint (soft parameter sharing), and (3) fully task-specific layers with independent parameters for each task. The full details of the model and dataset can be found in \ref{appendix:bikes}. The multi-task similarity computation took approximately four minutes on a workstation equipped with an Intel Core i7-8700 (6 cores/12 threads), 16 GB of RAM, an NVIDIA GeForce GTX 1060 (3 GB), and Ubuntu 24.04.2 LTS. 

\begin{figure}
\includegraphics[width =0.5\columnwidth]{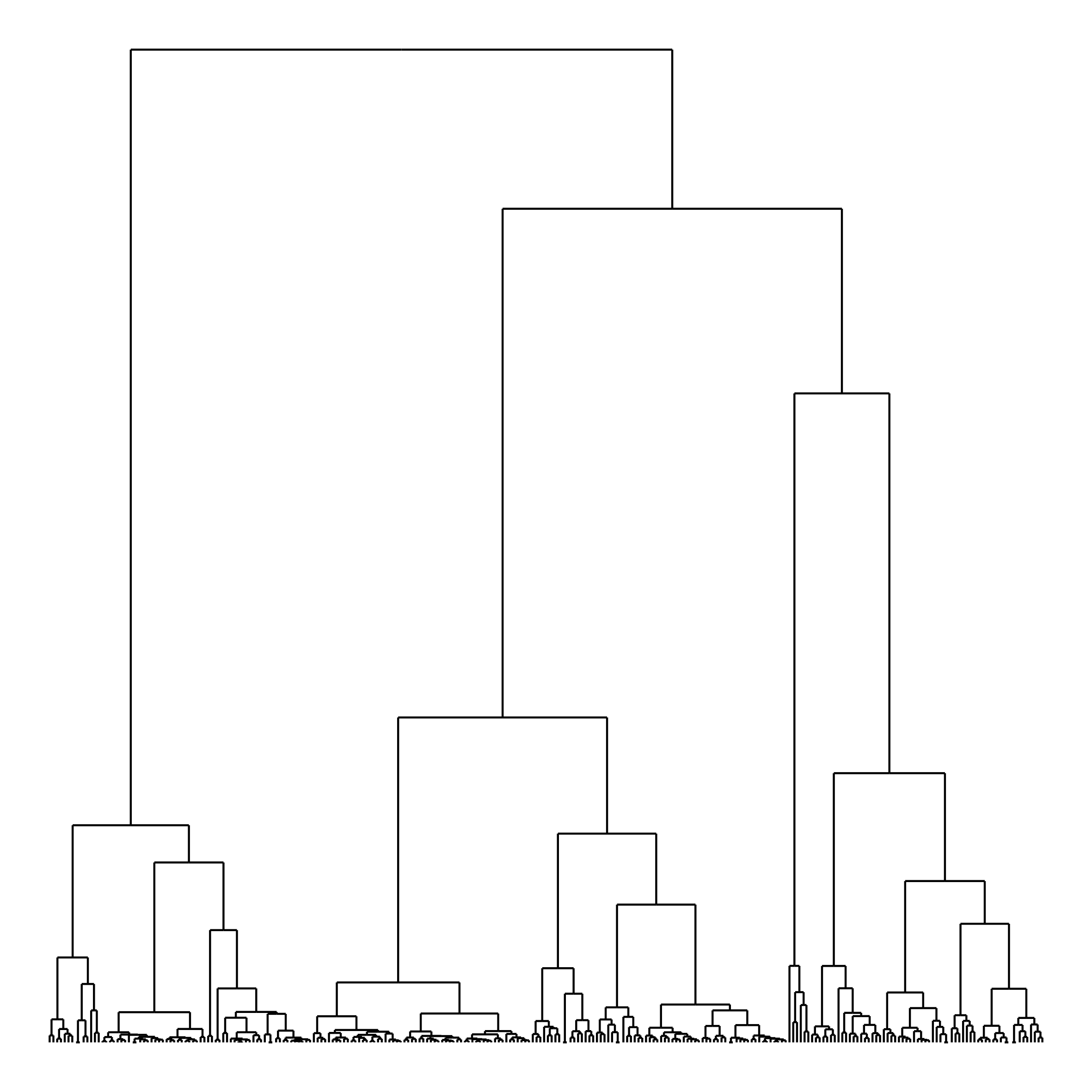}
\centering
\caption{Dendrogram for the hierarchical agglomerative clustering method using Ward’s linkage and the multi-task similarity measure values.}
\label{fig:dendro}
\end{figure}

The 264 tasks represent a relatively large number, making it challenging to manually inspect the relationships between them and derive a global interpretation of the model. This highlights the usefulness of the multi-task similarity measure in automatically extracting relationships between tasks. To explore and group tasks, the measure can act as a distance metric, enabling clustering algorithms to group similar tasks effectively. 

In this work, we use hierarchical clustering to group the tasks based on the multi-task similarity measure. Figure \ref{fig:dendro} presents the dendrogram obtained by applying a hierarchical agglomerative clustering method with Ward’s linkage. The dendrogram suggests the presence of six clusters. In this case, all features were considered equally important, so the similarity matrix is symmetric.

\begin{figure*}
\includegraphics[width=\textwidth]{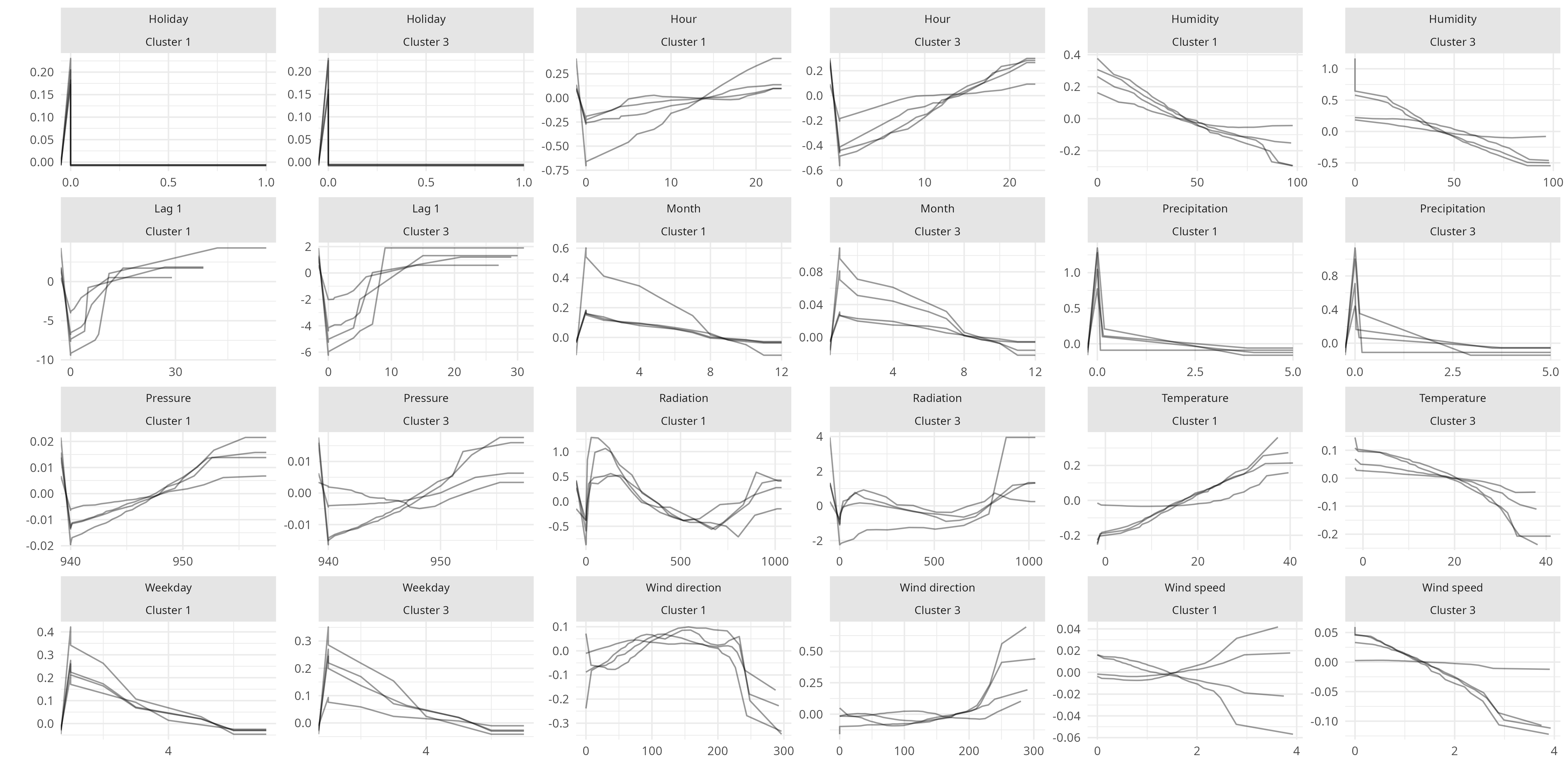}
\centering
\caption{Centered ALE plots for two of the clusters.}
\label{fig:ale-clusters-2-3}
\end{figure*}

Figure \ref{fig:ale-clusters-2-3} illustrates two of the resulting clusters. It can be observed that most features exhibit similar patterns in their ALE curves. However, there are three variables that present different behaviors. Specifically, the feature "Lag 1" shows a sharp positive effect in Cluster 1, while in Cluster 3 it initially increases but then exhibits a slight negative trend. Similarly, "Radiation" reveals a nonlinear relationship that peaks around mid-range values in Cluster 1, whereas in Cluster 3 it shows a steeper and more variable effect with a strong increase toward higher values. Lastly, "Temperature" has a consistently increasing effect in Cluster 1 but a clearly decreasing influence in Cluster 3. These differences suggest that while the overall model behavior is aligned across clusters, these features capture distinct functional dependencies, potentially reflecting different underlying subpopulation dynamics. This phenomenon, where most variables exhibit similar patterns and only a few show differences, highlights the importance of having measures like the one developed in this work.

\begin{figure*}
\includegraphics[width=\textwidth]{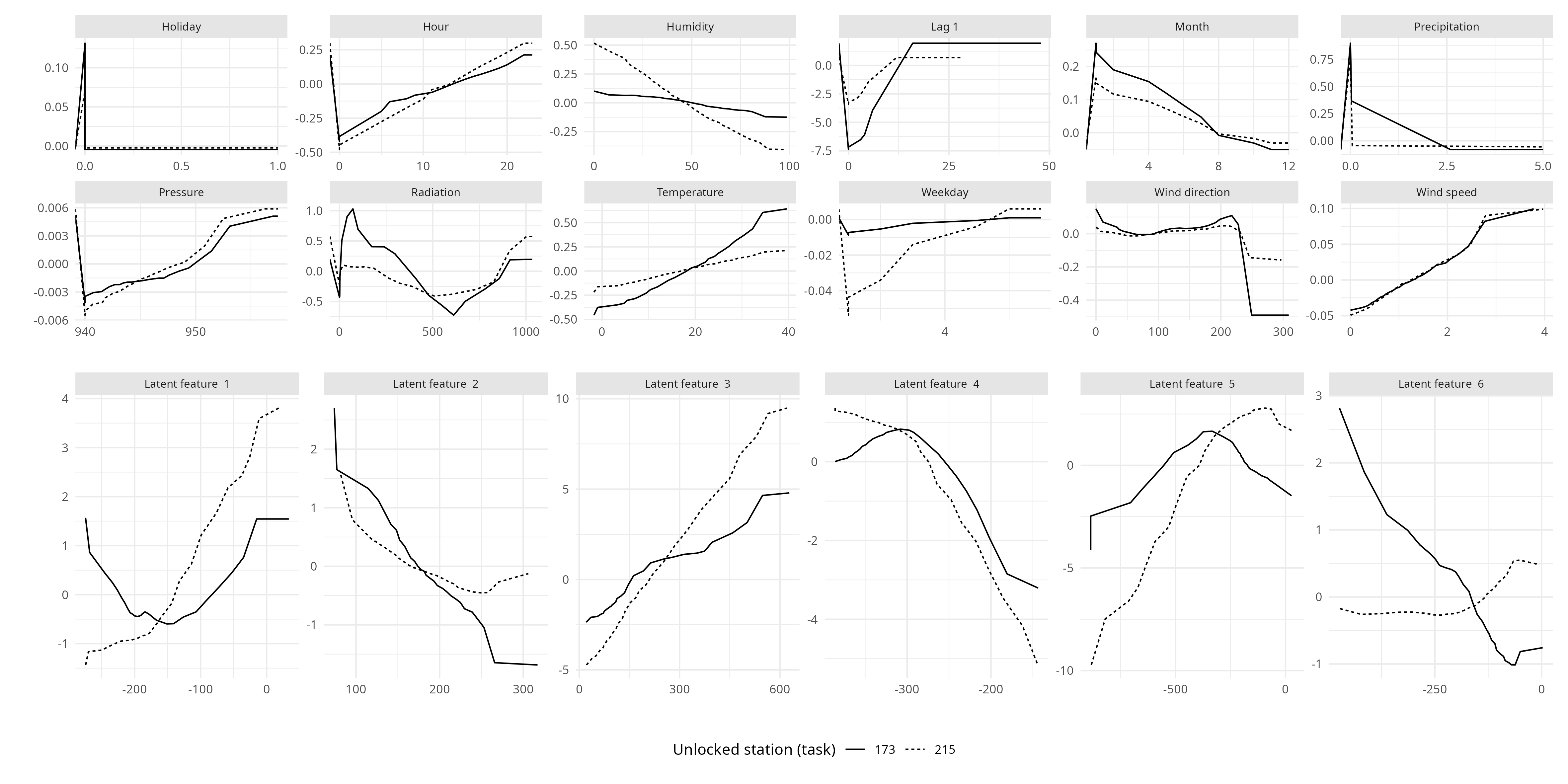}
\centering
\caption{Centered ALE plots for all features from the two most similar tasks.}
\label{fig:bike-most-sim}
\end{figure*}

\begin{figure*}
\includegraphics[width=\textwidth]{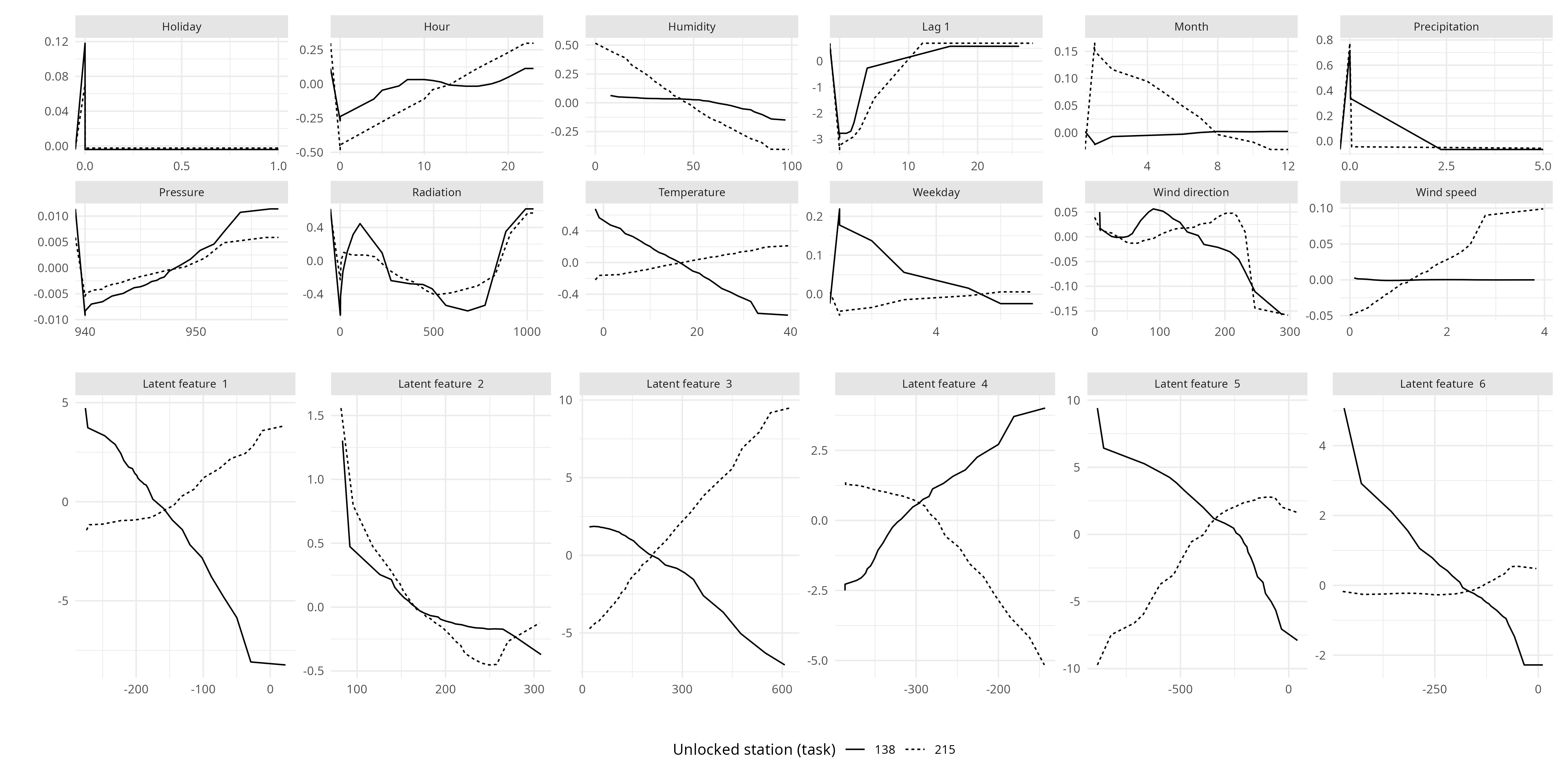}
\centering
\caption{Centered ALE plots for all features from the two most dissimilar tasks. The first twelve plots correspond to the original data, while the remaining six result from using the autoencoder.}
\label{fig:bike-most-dis}
\end{figure*}

Similar to the previous dataset, the differing patterns between the most similar and dissimilar tasks can be identified. The first twelve plots in Figures \ref{fig:bike-most-sim} and \ref{fig:bike-most-dis} show the ALE plots for the most similar and dissimilar tasks, respectively. Without a multitask similarity measure, it would not be straightforward to identify these tasks or to infer the nature of their differences.

The previous results, using a Random Forest model, can be found in~\ref{appendix:bikes}, showing that the outcomes are very similar. 

\subsubsection{Autoencoder}
\label{subsec:bicimad-autoencoder}

As discussed in Section~\ref{subsubsec:limitations}, the relationship between the input and output data can be sufficiently complex that it cannot be captured by first-order ALE curves. One possible remedy discussed in Section~\ref{subsubsec:limitations} is to use an autoencoder to learn a set of relevant latent features that better represent the underlying structure of the data. For example, in the bike sharing dataset, the effects of the hour of the day and the day of the week do not contribute independently to the prediction, because, as one can intuit, bike usage patterns differ significantly between weekdays and weekends depending on the time of the day. An autoencoder can help capture such interactions by encoding them into a more informative feature space and even discarding irrelevant features. The main drawback of this approach is that the interpretability of ALE curves applied to these new latent features may be reduced. Nevertheless, the representation can still serve as a useful similarity measure between tasks.

For this bike sharing dataset, we trained an autoencoder using all the data in the training set, without distinguish between tasks (i.e., stations unlocked in this context). The encoder outputs six latent features. These transformed features are then used as input to a separate model for each task. Further details are provided in the~\ref{appendix:bikes-autoencoder}.

\begin{figure*}
\includegraphics[width=\textwidth]{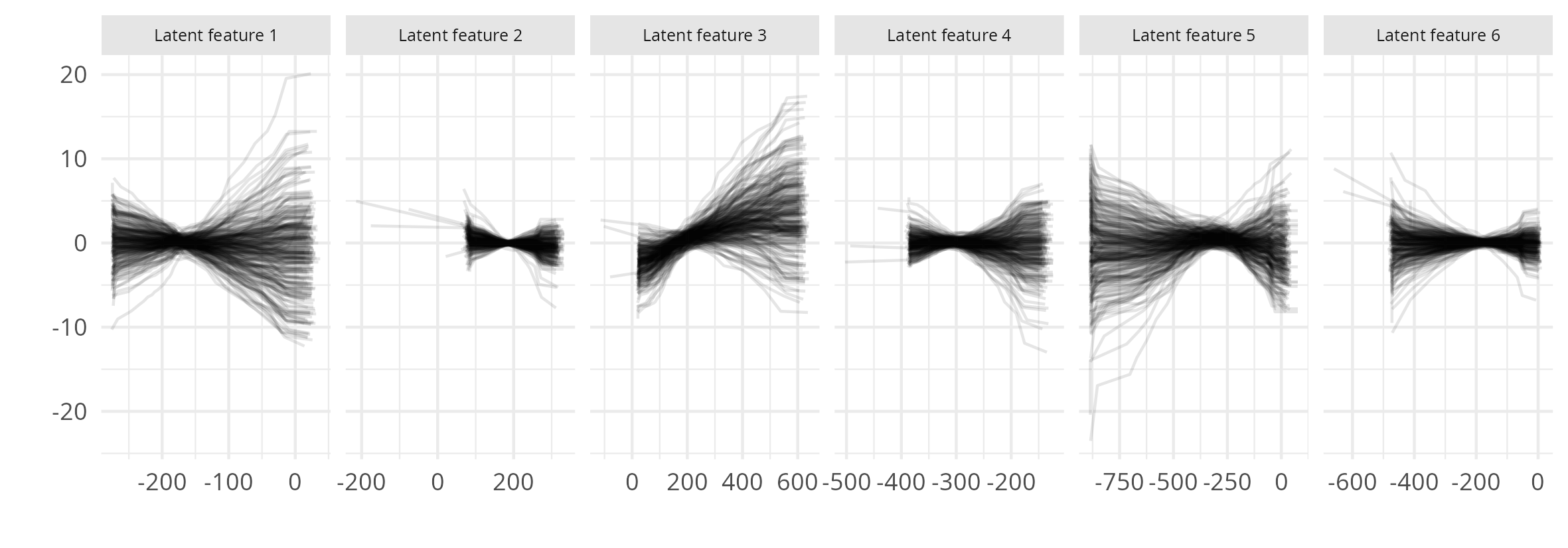}
\centering
\caption{Centered ALE plots for the four latent features obtained in the autoencoder.}
\label{fig:bikes-autoencoder-ales}
\end{figure*}

Figure~\ref{fig:bikes-autoencoder-ales} shows the ALE curves for each of the six latent features. Although these features lose the interpretability of the original input variables, the dimensionality reduction offers significant benefits—particularly in facilitating manual inspection of the relationships between inputs and outputs and, when possible, identifying some latent features as inherent properties of the data. The main advantage of the autoencoder-based representation is that complex relationships among the original features are captured and summarized within these six latent variables.

The last row of plots in Figures~\ref{fig:bike-most-sim} and~\ref{fig:bike-most-dis} presents the ALE curves for the same tasks previously discussed, but now using the autoencoder-based codification. Notably, the ALE plots exhibit much more distinguishable and structured patterns in the encoded latent features compared to those based on the original input variables.

\subsubsection{Spline smoothing}
\label{subsec:bicimad-spline}

\begin{figure*}
\includegraphics[width=\textwidth]{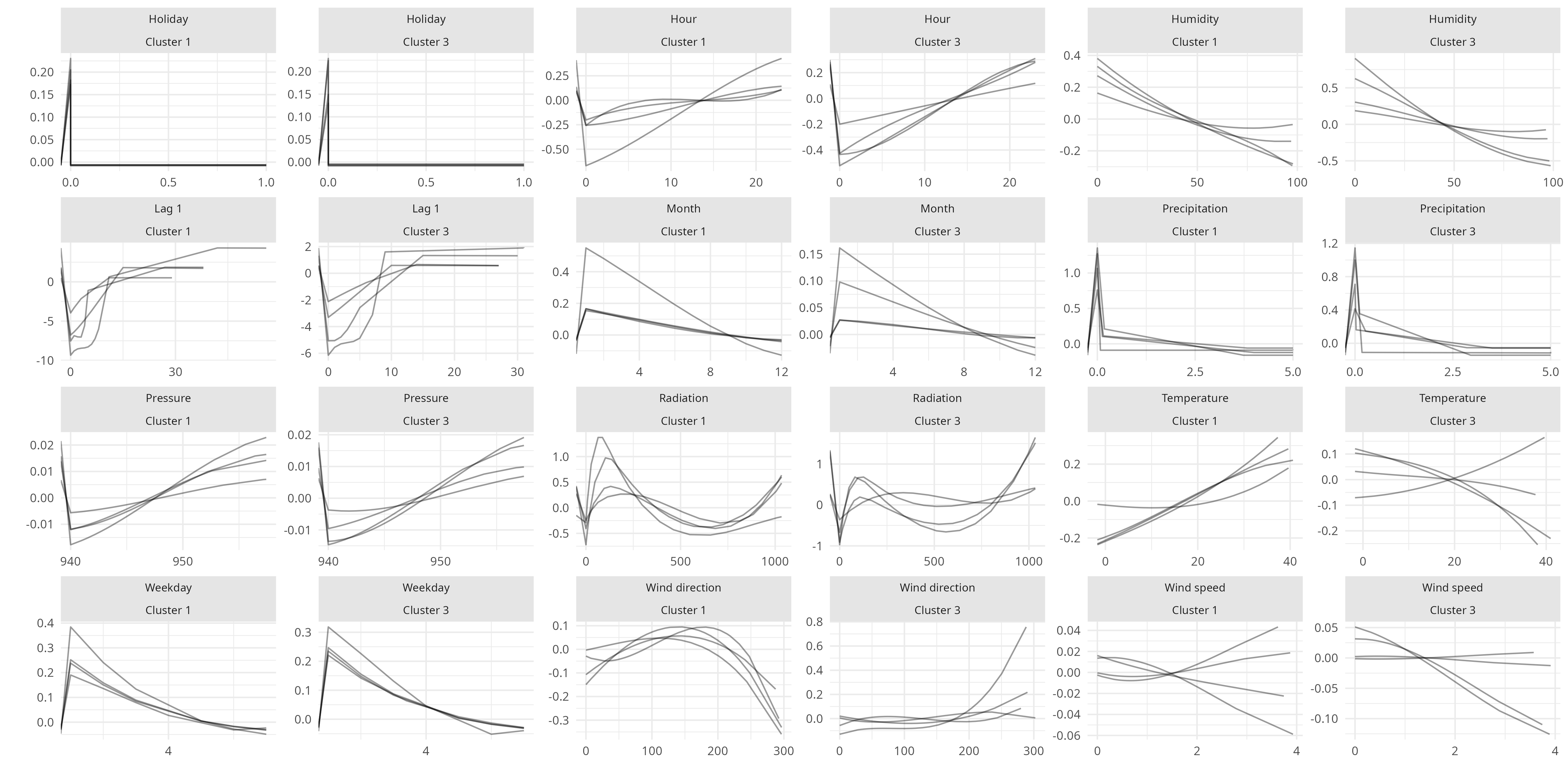}
\centering
\caption{Centered ALE plots for two of the clusters applying spline smoothing.}
\label{fig:ale-clusters-2-3-splines}
\end{figure*}

As discussed in Section~\ref{subsubsec:limitations}, the ALE curves can exhibit high-frequency oscillations that can obscure the interpretability of the method. Figure~\ref{fig:ale-clusters-2-3-splines} shows the same ALE curves as in Figure~\ref{fig:ale-clusters-2-3} but with roughness-penalized cubic spline smoothing applied. It can be observed that the rapid fluctuations present in the original curves are considerably attenuated, resulting in smoother and more interpretable profiles. While t
he overall trends of the effects are preserved, but the smoothed curves reduce visual noise, enabling for clearer identification of feature influence across clusters. 

\begin{table}[ht]
\centering
\caption{Comparison of multi-task similarity statistics with and without spline smoothing.}
\begin{tabular}{lcc}
\toprule
\textbf{Method} & \textbf{Mean Similarity} & \textbf{Standard Deviation} \\
\midrule
Without Smoothing & 23.682 & 13.281 \\
With Spline Smoothing & 26.523 & 13.342 \\
\midrule
Similarity differences  & -0.945 & 0.658 \\
\bottomrule
\end{tabular}
\label{tab:similarity-spline-comparison}
\end{table}

Although spline smoothing improves the interpretability of ALE curves by reducing high-frequency noise, it does not significantly alter the values of the multi-task similarity measure between tasks. Table~\ref{tab:similarity-spline-comparison} shows that both the mean and standard deviation of similarity values remain relatively stable before and after applying spline smoothing. The mean similarity increases slightly from 23.682 to 26.523, and the standard deviation remains nearly unchanged. Furthermore, the root mean square error (RMSE) of the similarity values between the two configurations is $1.151$, indicating minimal overall deviation introduced by the smoothing process. Additionally, a Spearman correlation of $0.9953$ is obtained.

These results support the robustness of the proposed similarity measure with respect to minor variations in curve smoothness.

\subsection{Image dataset}
\label{subsec:img}

The previous datasets share the common characteristic of being tabular, meaning the information is structured as observations and variables in a table-like format. While this structure is naturally well-suited for use with the multitask similarity measure, the measure can also be applied to other data types, provided the information is appropriately structured.

In Section~\ref{subsubsec:limitations}, we proposed several strategies to apply the measure to non-tabular data. Although the approaches differ in how the latent representation is computed, they all share the same underlying idea: encoding the original data into a latent space.

To provide experimental validation of these ideas, we applied the concept bottleneck encoder to the CelebA image dataset~\footnote{\url{https://mmlab.ie.cuhk.edu.hk/projects/CelebA.html}}. This dataset contains facial images annotated with multiple binary attributes, which serve as human-interpretable concepts and are well suited for evaluating concept-based representations. Each image is annotated with 40 binary attribute such as \textit{Smiling, Wearing Glasses}, or \textit{Mustache}. It is reasonable to assume that some of these attributes are correlated, for instance, images labeled as \textit{Male} are likely to have a higher probability of also being labeled as \textit{Mustache}.

In our setup, half of the attributes are used as concepts in the concept bottleneck encoder, while the other half serve as separate outputs, each representing an individual task in the multitask learning setting. Further details can be found in~\ref{appendix:celeba}.

The neural network employed consisted of a Convolutional Neural Network (CNN) used as an encoder to extract a latent concept representation from the input images. This encoder maps each image to a 20-dimensional binary concept vector through a series of convolutional and fully connected layers, using ReLU activations in the hidden layers and sigmoid activations at the output to produce probabilities for each concept.

These predicted concept representations serve as input to a multitask classifier, implemented as a fully connected neural network that independently predicts the 20 target attributes ---each corresponding to a separate binary classification task. The full model is trained end-to-end using a combined loss function that includes binary cross-entropy for both concept prediction and final multitask outputs, thereby ensuring the consistency and relevance of the latent representations.

Once trained, the multitask similarity measure is applied to the ALE curves computed for each concept. This allows us to evaluate how variations in the predicted probability of each concept affect the probability of the output attributes across tasks. By quantifying the similarity between tasks in terms of these concept-based ALE curves, we can better understand the functional relationships learned by the model and the role of shared or divergent concept influences.

Table~\ref{tab:celeba-task-similarity} presents the pairwise similarity between four representative target task attributes from the CelebA dataset, as computed by the proposed multitask similarity measure based on ALE curves over the learned concept space. The pairwise distances confirm several intuitive patterns.
The \emph{Mustache} and \emph{Male} tasks form the closest pair, with a similarity value of only $0.17$, reflecting the strong association between facial hair and gender in CelebA.
At the other extreme, \emph{Mustache} and \emph{Wearing Lipstick} are separated by the largest value ($0.88$), underscoring that these tasks depend on markedly different concept cues. A cosmetically driven relationship is evident between \emph{Wearing Lipstick} and \emph{Wearing Necklace} ($0.36$), the next-smallest distance in the matrix, suggesting that similar latent features—likely linked to femininity and facial accessories—govern both outputs.

Finally, intermediate values ($0.53 - 0.73$) for pairs involving \emph{Male} and \emph{Wearing Necklace} signal a moderate overlap of underlying concepts, whereas distances above $0.80$ denote tasks with largely distinct functional dependencies. Overall, lower distances consistently correspond to semantically plausible task similarities, validating the usefulness of the ALE-based measure.

Table~\ref{table:celeba-similar-tasks-var} confirms that the multi-task similarity measure capture semantically meaningful relations between concepts and tasks. Gender-linked cues—especially \textit{Bald}, \textit{Black Hair}, and \textit{Eyeglasses}—exhibit near-zero distances to \textit{Male} ($0.01$, $0.02$, and $0.03$, respectively) and similarly small values to \textit{Mustache}, reflecting their shared dependence on male identity. In contrast, the cosmetic concept \textit{Heavy Makeup} aligns most strongly with the femininity cue \textit{Wearing Lipstick} ($0.05$, the lowest entry in its row), while all concepts show markedly larger distances to the accessory task \textit{Wearing Necklace} ($0.11–0.41$). Hence lower distances consistently correspond to intuitive task–concept affinities, demonstrating that the multitask-similarity metric surfaces the expected gender and cosmetic patterns while down-weighting weakly informative features.

\begin{table}[htbp]
\footnotesize
\centering
\caption{Pairwise similarity between selected output tasks in the CelebA dataset based on ALE curve comparisons over concept representations. Lower values indicate more similar tasks.}
\label{tab:celeba-task-similarity}
\begin{tabular}{lcccc}
\toprule
\textbf{Task} & \textbf{Mustache} & \textbf{Male} & \textbf{Wearing Lipstick} & \textbf{Wearing Necklace} \\
\midrule
\textbf{Mustache}          & --  & 0.17 & 0.88 & 0.73 \\
\textbf{Male}              & 0.17  & -- & 0.81 & 0.53 \\
\textbf{Wearing Lipstick}  & 0.88  & 0.81 & -- & 0.36 \\
\textbf{Wearing Necklace}  & 0.73  & 0.53 & 0.36 & -- \\
\bottomrule
\end{tabular}
\end{table}

\begin{table}[htbp]
\small
\caption{Multi–task similarity measure between concepts and two CelebA tasks.  
Each entry is the weighted Fr\'echet distance (lower $\Rightarrow$ stronger relation).  }
\centering
\begin{tabular}[t]{rcccc}
\toprule
 & Mustache & Male & Wearing Lipstick & Wearing Necklace\\
\midrule
\multicolumn{5}{l}{\textbf{Mustache}}\\
\hspace{1em}Bald        & -- & \textbf{0.01} & 0.11 & 0.12\\
\hspace{1em}Heavy Makeup   & -- & 0.05 & \textbf{0.05} & 0.07\\
\hspace{1em}Black Hair     & -- & \textbf{0.02} & 0.17 & 0.13\\
\hspace{1em}Eyeglasses     & -- & \textbf{0.03} & 0.13 & 0.11\\
...     &  &  ...& ...& ...\\
\midrule
\multicolumn{5}{l}{\textbf{Male}}\\
\hspace{1em}Bald        & \textbf{0.01} & -- & 0.51 & 0.33\\
\hspace{1em}Heavy Makeup   & 0.05 & -- & \textbf{0.80} & 0.28\\
\hspace{1em}Black Hair     & \textbf{0.02} & -- & 0.13 & 0.37\\
\hspace{1em}Eyeglasses     & \textbf{0.03} & -- & 0.12 & 0.41\\
...     & ... &  & ...&... \\
\bottomrule
\end{tabular}
\label{table:celeba-similar-tasks-var}
\end{table}

\section{Results and Discussion}
\label{sec:results-discussion}

The application of the multi-task similarity measure across the datasets discussed in Section~\ref{sec:empiricalWork} highlights its utility in identifying and explaining relationships between tasks. The measure’s outcomes align with intuitive expectations, enabling the identification of tasks with comparable relationships between features and outcomes from an explainable perspective.

The synthetic dataset of Section~\ref{subsec:syntheticdata1} serves an ideal testbed to validate the measure in a controlled environment. The computed similarity values correspond with visual observations of the ALE curves, underscoring the measure's reliability. In this ad hoc dataset, consisting of only five tasks with five variables each, it is straightforward to manually identify similar tasks without requiring a formal measure. Nevertheless, the computed similarity values align with these manual observations, further validating the measure.

Conversely, the relatively small Parkinson's dataset, comprising 42 patients with approximately 200 records per patient, presents a significant challenge in identifying task similarities (patients, in this context) without an appropriate similarity measure. This underscores the importance of the multi-task similarity measure in uncovering such relationships and facilitating the derivation of meaningful conclusions. 

Furthermore, the third dataset, used for predicting bike-sharing usage, involves 264 tasks. Identifying task dynamics in this scenario without a summarization tool is particularly challenging. Here, the similarity measure can serve as the foundation for clustering methods to group tasks effectively. Importantly, the measure captures both the overall similarity and variable-specific influences, validating its robustness and interpretability across datasets of varying complexity.

A logical sequence in utilizing the measure involves first computing the multi-task similarity measure between all the tasks, or at least between one task of interest and the remaining tasks. Second, it is beneficial to examine not only the most similar tasks but also the least similar ones. By preserving intermediate calculations necessary for computing the measure, it becomes possible to elucidate why and how tasks are similar or dissimilar. This includes identifying the relevant variables that contribute to the low or high values of the similarity measure.

Additionally, the measure allows for intervention in its behavior to emphasize certain variables. Ideally, the importance of each variable used in the similarity calculation should be determined using a reliable method. However, researchers can manually adjust the variable weights to prioritize those deemed most relevant or actionable for a specific study.

Several multi-task learning scenarios presuppose, either implicitly or explicitly, a degree of homogeneity among the tasks. The proposed measure is robust enough to be applied to heterogeneous tasks, even when the relationships are hidden behind task structures. For instance, if one task's data includes a variable called \textit{age} and another task's data includes \textit{edad} (age in Spanish), the measure can detect the similarity between these variables despite the difference in nomenclature, provided they represent the same information.

A significant advantage of this measure is its model-agnostic nature. It is irrelevant which model has been used to train each task, and it can even handle a mix of models. For example, some tasks may be trained using black-box machine learning models like deep learning, while others use traditional statistical models like linear regression. This is because the measure compares ALE curves derived from the models, and ALE curves are inherently model-agnostic.

However, the validity of the results heavily depends on the quality of the trained models. If the model for each task is not sufficiently capable of capturing the true patterns in the data and the relationships between predictor variables and the target, the resulting measure may not be accurate. Conclusions and subsequent actions are limited if the models are not well-developed. It is critical to ensure that all models used meet appropriate quality standards and exhibit suitable behavior. Furthermore, the measure can serve as an exploratory tool to identify defective models, allowing researchers to refine and improve them.

One challenge in using the measure lies in how the ALE curves are computed. Since the Fr\'echet distance defined here uses a summation instead of the maximum, significant differences in the number of segments in the ALE curves between tasks can affect the reliability of the measure's results. This issue can be alleviated if all ALE curves have the same number of segments, which makes the measure more robust. At the very least, if it is known which variables to be compared, it is recommended that all of the ALE curves corresponding to these variables have the same number of segments.

Like any machine learning technique, the multi-task similarity measure can help achieve its defined purpose, but it must be used cautiously and not applied uncritically.

\section{Conclusions and Future Work}
\label{sec:conclusions}

In this work, we explored the construction of a multi-task similarity measure from an explainable perspective. This approach aims to identify similar tasks in a multi-task scenario and provide insights into why and how they are similar. While several similarity measures exist for comparing tasks, particularly in multi-task learning, --- none explicitly facilitate the explanation of task similarities--- which is invaluable for researchers.

Throughout the paper, we applied the measure to four datasets: one synthetic dataset designed to demonstrate the behavior of the measure in a controlled environment, a real dataset of Parkinson’s patients, a dataset focused on predicting bike-sharing usage, and an image dataset (CelebA) using concept bottleneck models. Our results demonstrate the measure's utility in detecting task similarities, highlighting the intuitive understanding of similarity provided by the multi-task similarity measure. 

Furthermore, although there are contexts in which the number of tasks may pose computational challenges, the execution times for the relatively large Bike Sharing Dataset (264 tasks) are reasonable and manageable even on modest hardware.

This research opens up several avenues for future work. Firstly, as demonstrated with the bike sharing dataset, the similarity measure can be integrated into clustering methods to automatically group similar tasks. An important line of research would involve identifying which clustering methods are most suitable for this purpose, considering factors such as scalability, interpretability, and robustness. Additionally, the quality of the measure can be studied across different models with diverse underlying hypotheses (e.g., linear versus non-linear models).

Secondly, the similarity measure can be incorporated into multi-task learning algorithms. For instance, in soft parameter-sharing algorithms, the constraints on shared parameters could be derived directly from the similarity measure. This would allow the algorithm to dynamically adjust parameter sharing based on the quantified similarity between tasks, potentially improving both performance and interpretability.


\section*{Acknowledgements}
The authors thank their respective universities for the support and Daniel also acknowledges the project PID2021-125645OB-I00 (PARCHE), funded by MCIN/AEI/10.13039/501100011033/FEDER, EU.

\bibliographystyle{elsarticle-num} 

\bibliography{references}


\appendix
\section{Synthetic Dataset 1}
\label{appendix:dataset-1}
We simulate five tasks, each with $10,000$ observations and five features generated as follows:
\begin{itemize}
    \item $X_1$ and $X_2$ are simulated from a bivariate normal distribution with the same mean $\mu$ and covariance matrix $\Sigma$ across all tasks,
    \item $X_3$ is simulated uniformly in the $(0, 1)$ interval, and
    \item  $X_4$ and $X_5$ are simulated as a mixture of normals, with each task having its own parameters.
\end{itemize}

The outcome $Y$ is generated as shown in Equation \ref{eq:synth1}.

\begin{equation}
\label{eq:synth1}
    Y = Rastrigin_{std}(X_1, X_2) + q_{std}(X_4, X_5).
\end{equation}

The first part of the sum is computed as

\begin{equation}
\label{eq:rastrigin}
    Rastrigin(X_1, X_2) = 20 + \sum_{i=1}^2(X_i^2-10\cos(2\pi X_i))
\end{equation}

The second part is the quadratic form
\begin{equation}
    q(X_4, X_5) = aX_4^2+ bX_5^2 + cX_4X_5
\end{equation}
\noindent where the scalars $a,b,c\in\{-1,1\}$ are dependent of the task.

The functions $Rastrigin_{std}(X_1, X_2)$ and $q_{std}(X_4,X_5)$ appearing in Equation~\ref{eq:synth1} are the normalized versions of the respective functions, i.e., with zero-mean and unit variance. For a detailed breakdown of the simulation, we refer to Table~\ref{table:synth1-params}.

\begin{table*}[!t]
\centering
\caption{Simulated data generation for each of the five tasks. $\mathcal{N}_2(\mu, \Sigma)$ refers to the bivariate normal of mean $\mu$ and standard deviation $\Sigma$, $\mathcal{U}(a, b)$ refers to uniform distribution in the interval $(a,b)$ and $\mathcal{N}(\mu_1, \Sigma_1) + \mathcal{N}(\mu_2, \Sigma_2)$ is a mixture of two normals. The $a$, $b$ and $c$ are the parameters of the quadratic form expressed in Equation~\ref{eq:synth1}.}
\begin{tabular}{r||cc|c|cccc|cccc|cccc}
\toprule
 & \multicolumn{2}{c|}{$X_1$ and $X_2$} & $X_3$ & \multicolumn{4}{c|}{$X_4$ and $X_5$}  & \multicolumn{3}{c}{$Y$} \\
\midrule
\textbf{Task} & $\mathcal{N}_2(\mu$ & $\Sigma)$ & $\mathcal{U}(a,b)$ &$\mathcal{N}(\mu_1$&$\Sigma_1)\,+$&$\mathcal{N}(\mu_2$&$\Sigma_2)$&a&b&c\\
\hline
\hline
\textbf{1} &  & & & 0 & 0.1 & 0 & 0.1 & 1 & 1 & 1 \\  
\textbf{2} &  & & & 0 & 0.1 & 0 & 0.1 & 1 & 1 & -1 \\  
\textbf{3} & $(0,0)$ & $\begin{pmatrix}
2 & 1\\
1 & 2
\end{pmatrix}$ &$(0,1)$& -0.25 & 0.1 & 0.25 & 0.1 & 1 & -1 & 1 \\ 
\textbf{4} &  & & & 0 & 0.1 & 0 & 0.1 & -1 & 1 & 1 \\  
\textbf{5} &  & & & -0.25 & 0.1 & 0.25 & 0.1 & -1 & -1 & 1 \\  
\bottomrule
\end{tabular}
\label{table:synth1-params}
\end{table*}

The relationship of the predictors with the outcome, as per Equation~\ref{eq:synth1}, can be decomposed into two additive components. 

\begin{figure*}
    \centering
    \begin{subfigure}[t]{.7\textwidth}
        \centering
        \includegraphics[width = .45\textwidth]{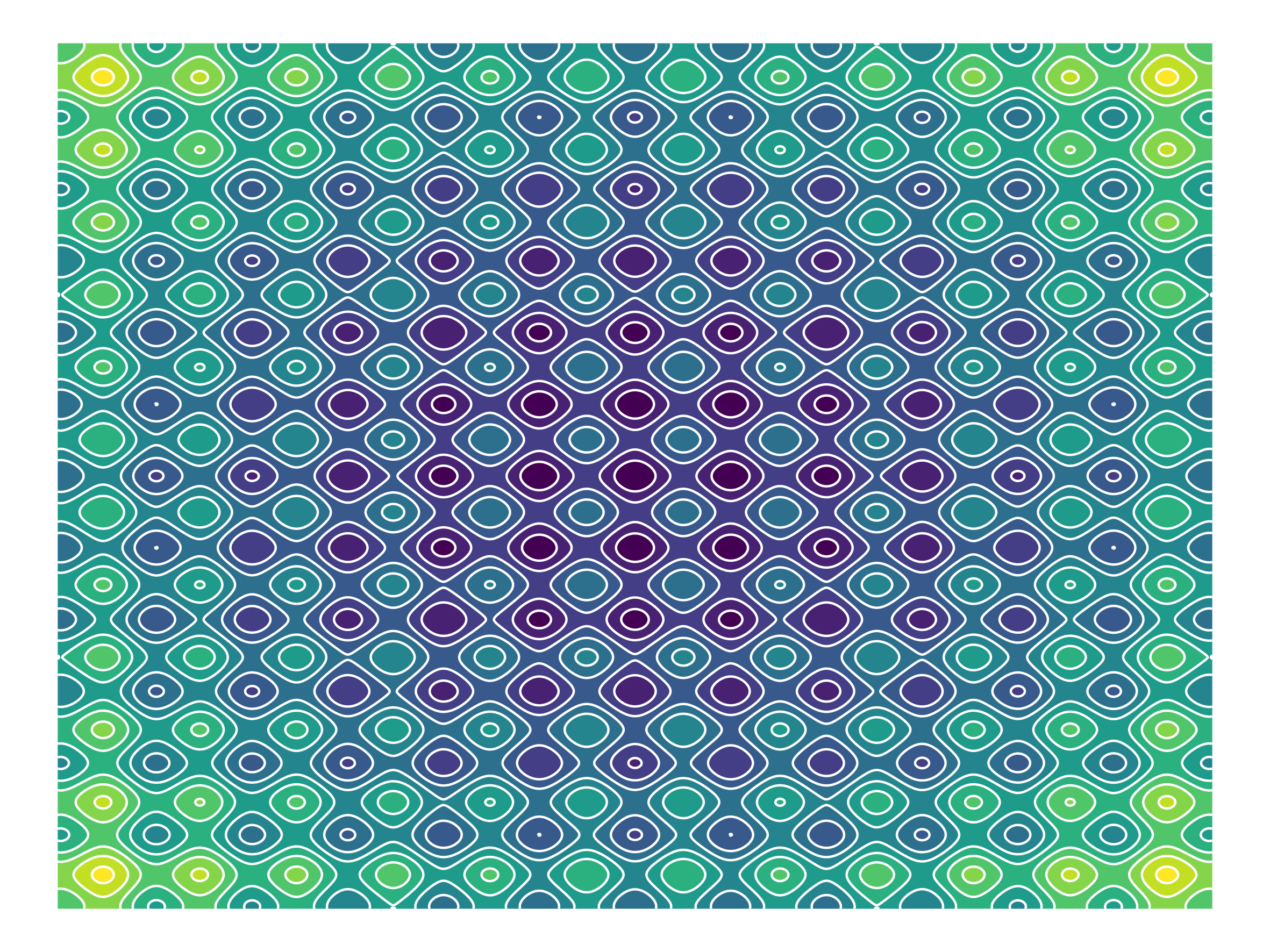}
        \includegraphics[width = .45\textwidth]{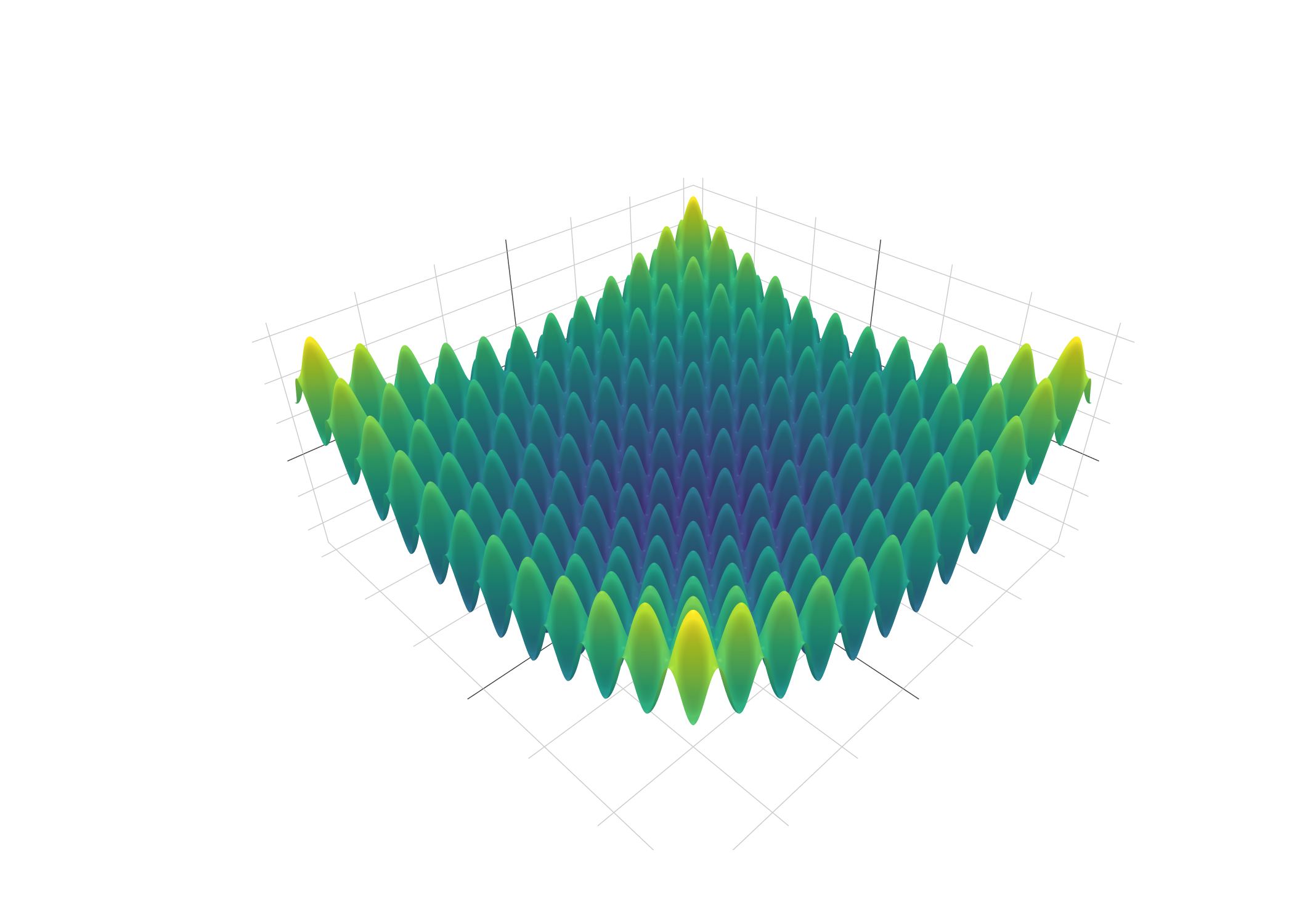}
    \end{subfigure}
\caption{Surface plots and contour plots of the relationship between predictor variables $X_1$ and $X_2$ and the outcome across all tasks.}
\label{fig:synth1-rastrigin}
\end{figure*}

The first one is attributed to $X_1$ and $X_2$ through the $Rastrigin$ function described in Equation~\ref{eq:rastrigin}. This complex function is commonly employed to test the performance of optimization algorithms, such as genetic algorithms~\cite{muhlenbein_parallel_1991}. It is characterized by its complexity, featuring numerous local minima and a large search space. The function's graphical representation can be observed in Figure~\ref{fig:synth1-rastrigin}. 

\begin{figure*}
    \begin{subfigure}[t]{.47\textwidth}
        \centering
        \includegraphics[width = .47\textwidth]{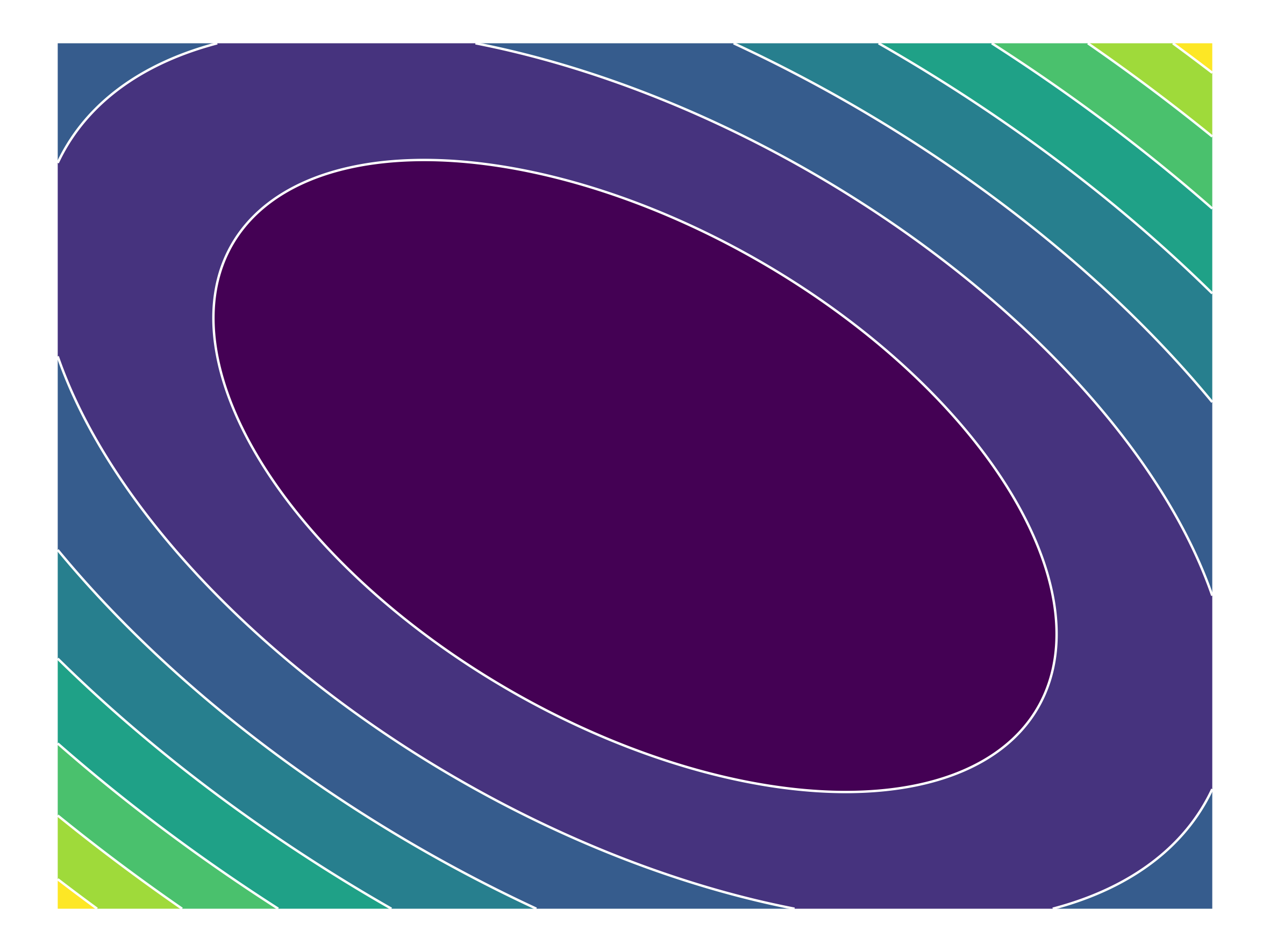}
        \includegraphics[width = .47\textwidth]{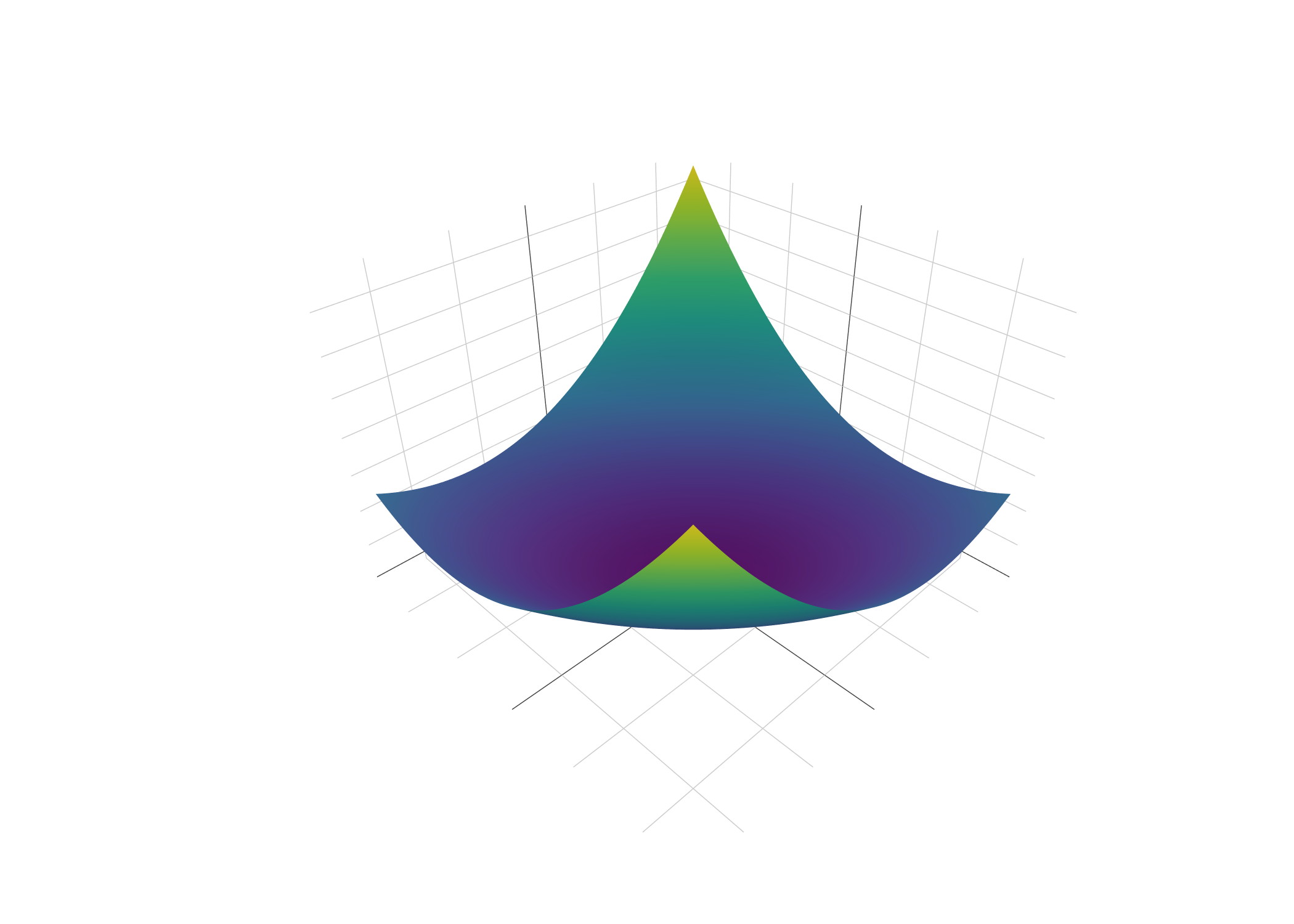}
        \caption{Task 1}
    \end{subfigure}
    \begin{subfigure}[t]{.47\textwidth}
        \centering
        \includegraphics[width = .47\textwidth]{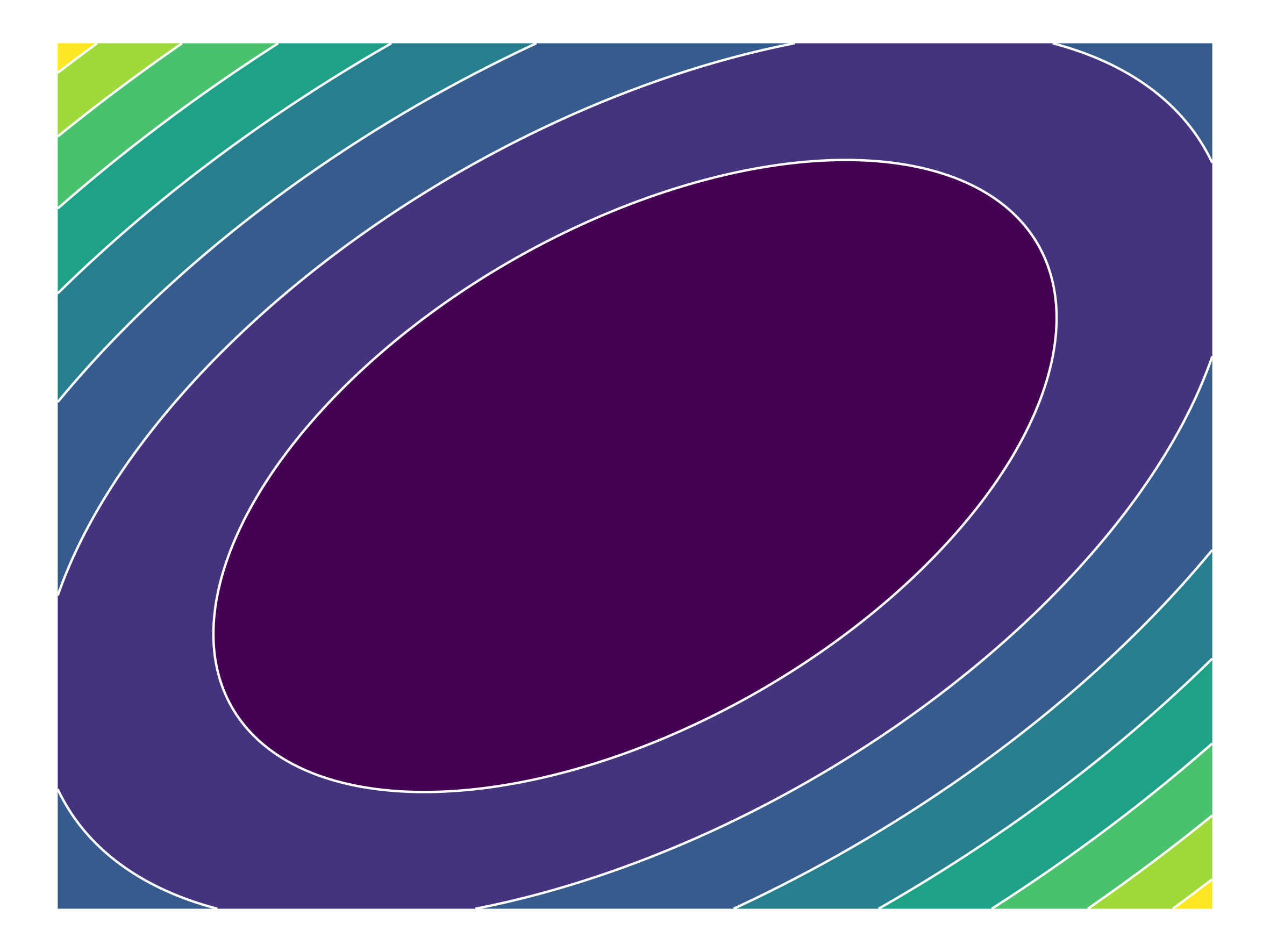}
        \includegraphics[width = .47\textwidth]{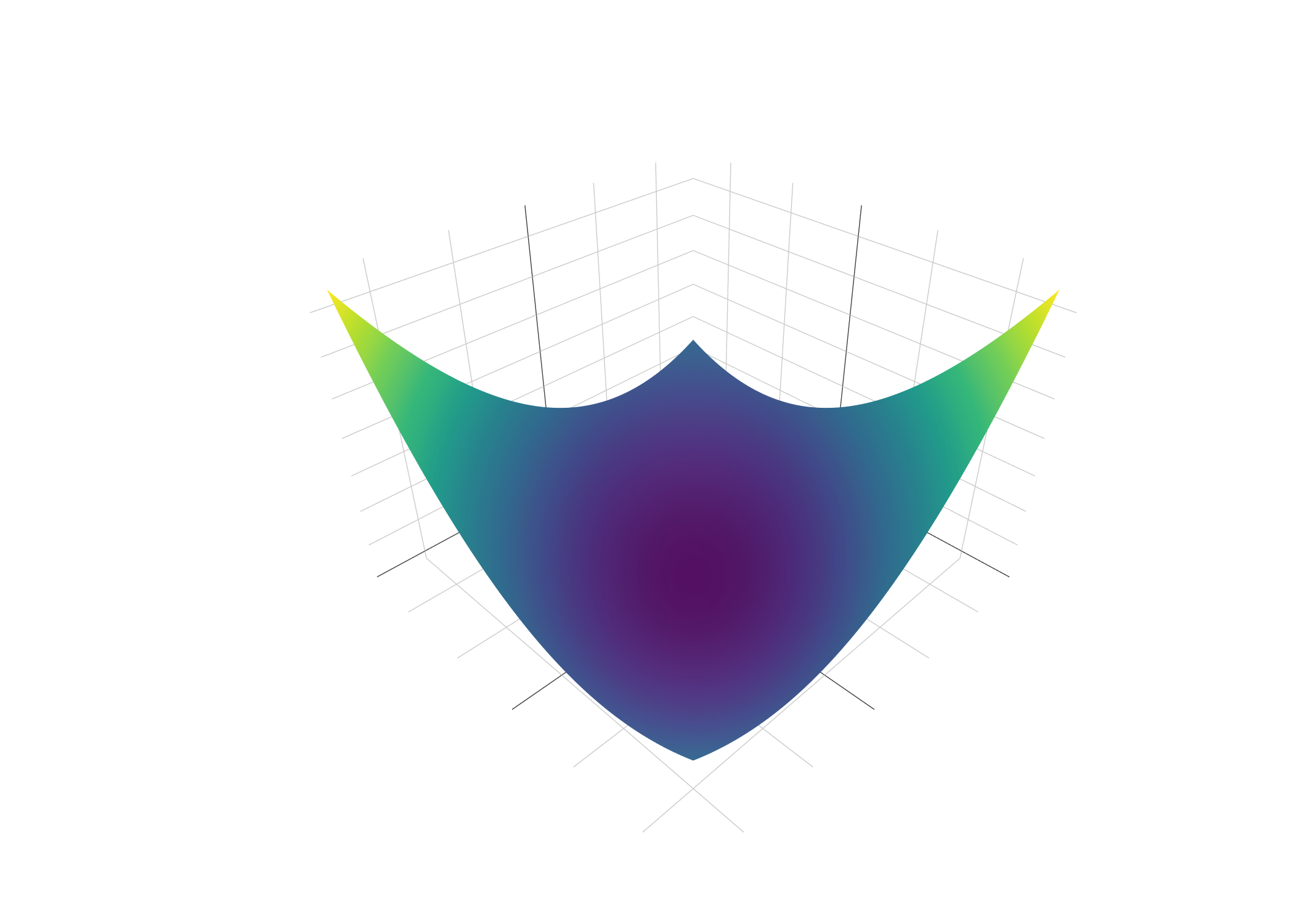}
        \caption{Task 2}
    \end{subfigure}
    \begin{subfigure}[t]{.47\textwidth}
        \centering
        \includegraphics[width = .47\textwidth]{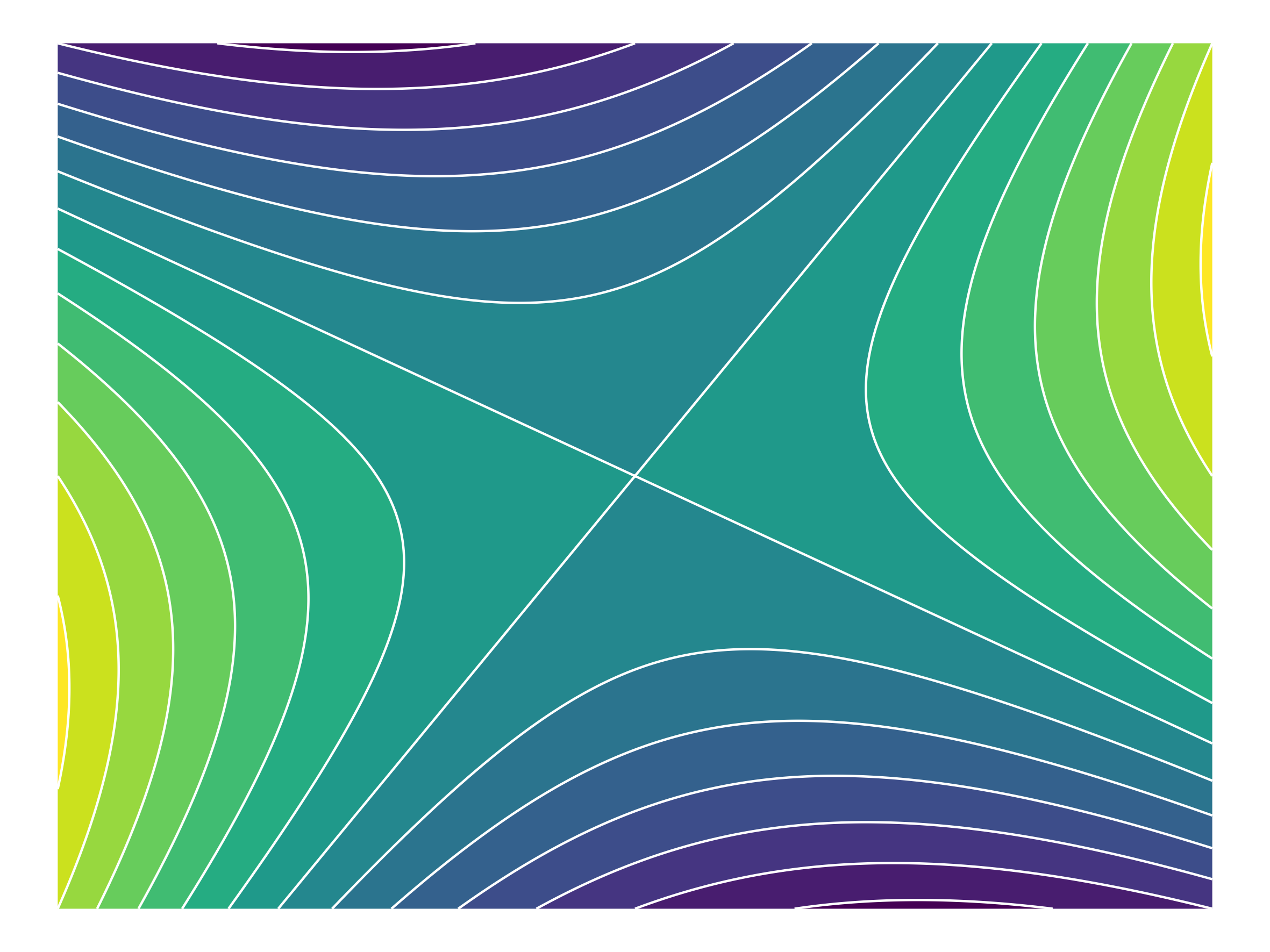}
        \includegraphics[width = .47\textwidth]{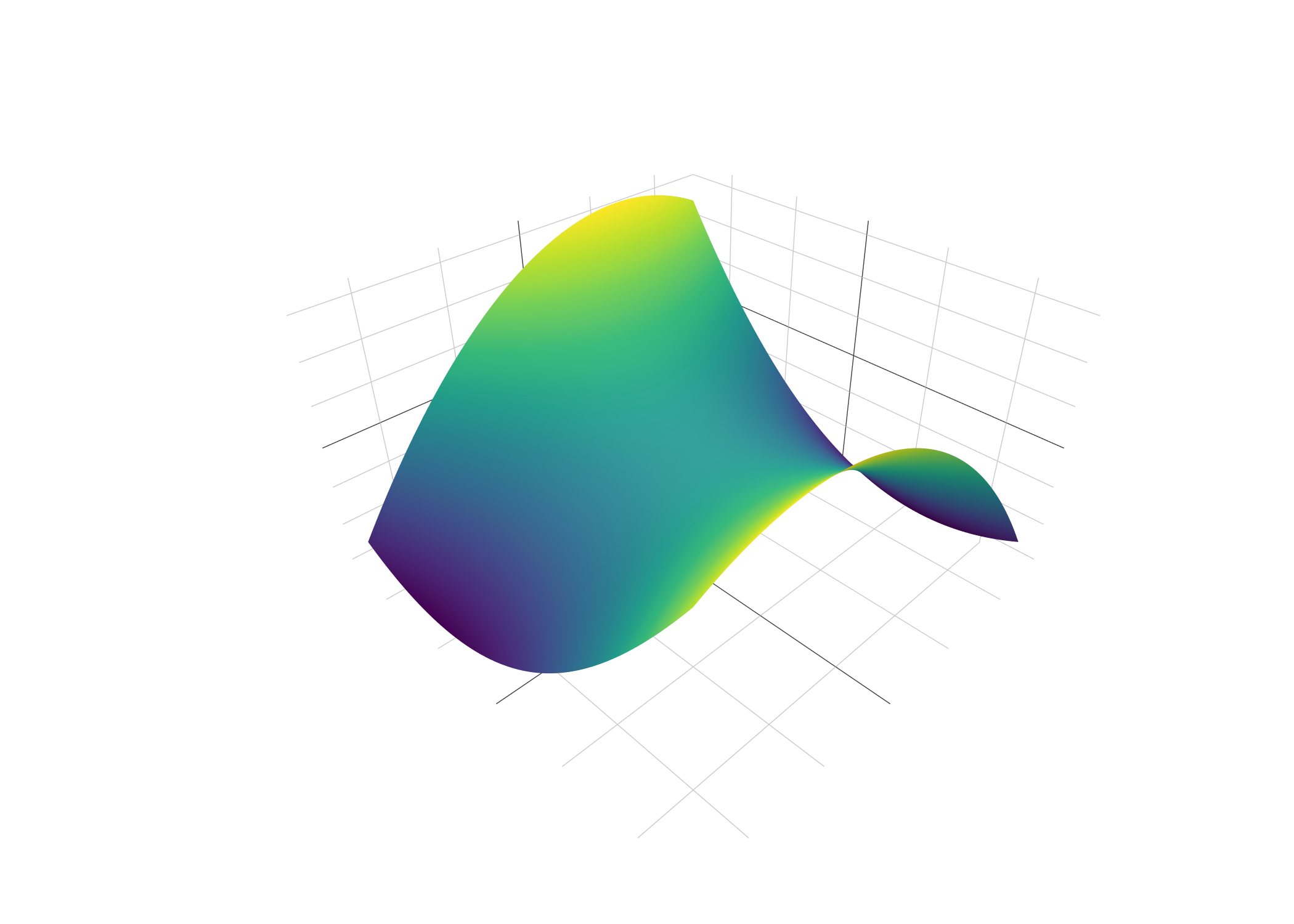}
        \caption{Task 3}
    \end{subfigure}
    \begin{subfigure}[t]{.47\textwidth}
        \centering
        \includegraphics[width = .47\textwidth]{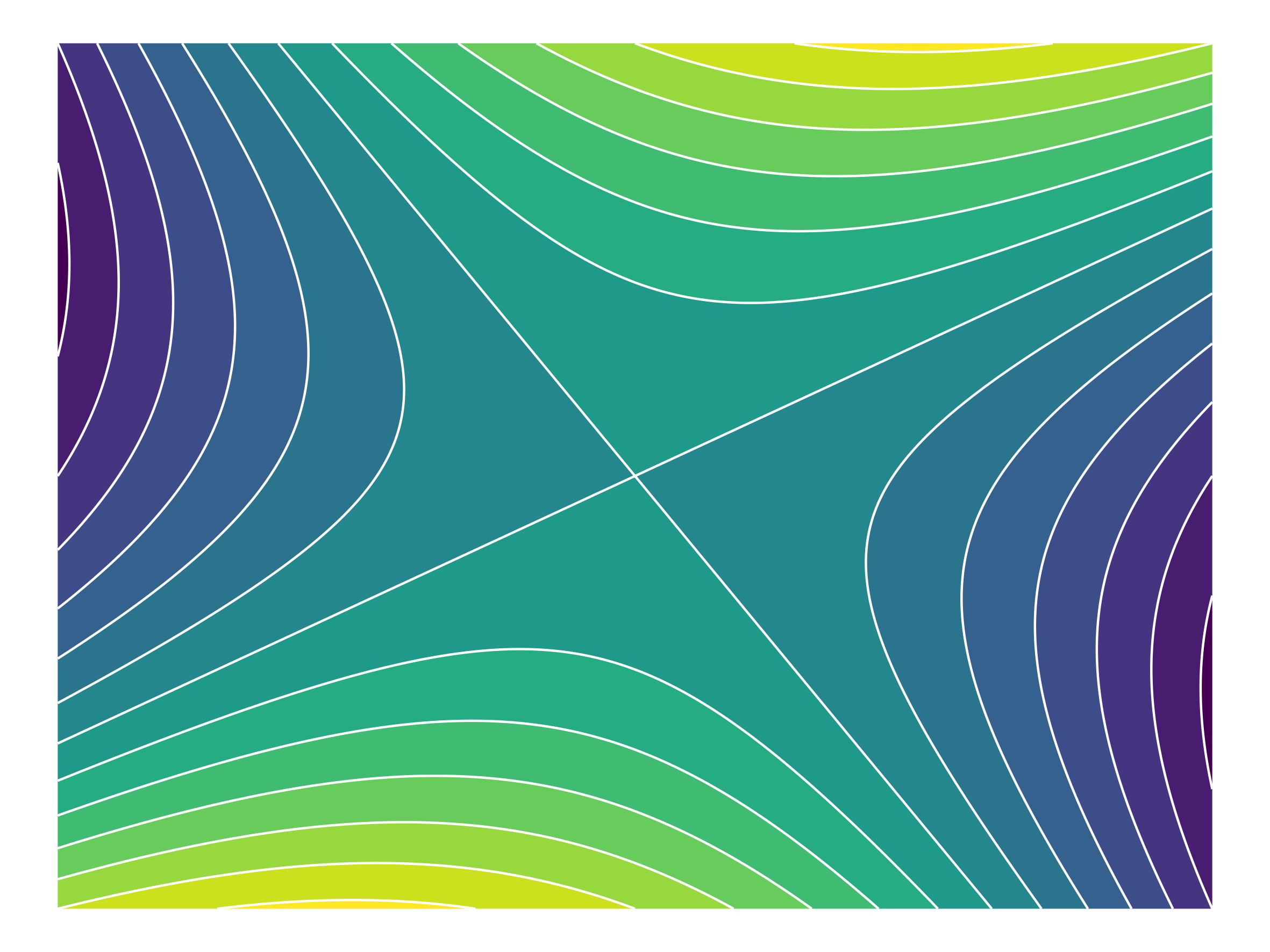}
        \includegraphics[width = .47\textwidth]{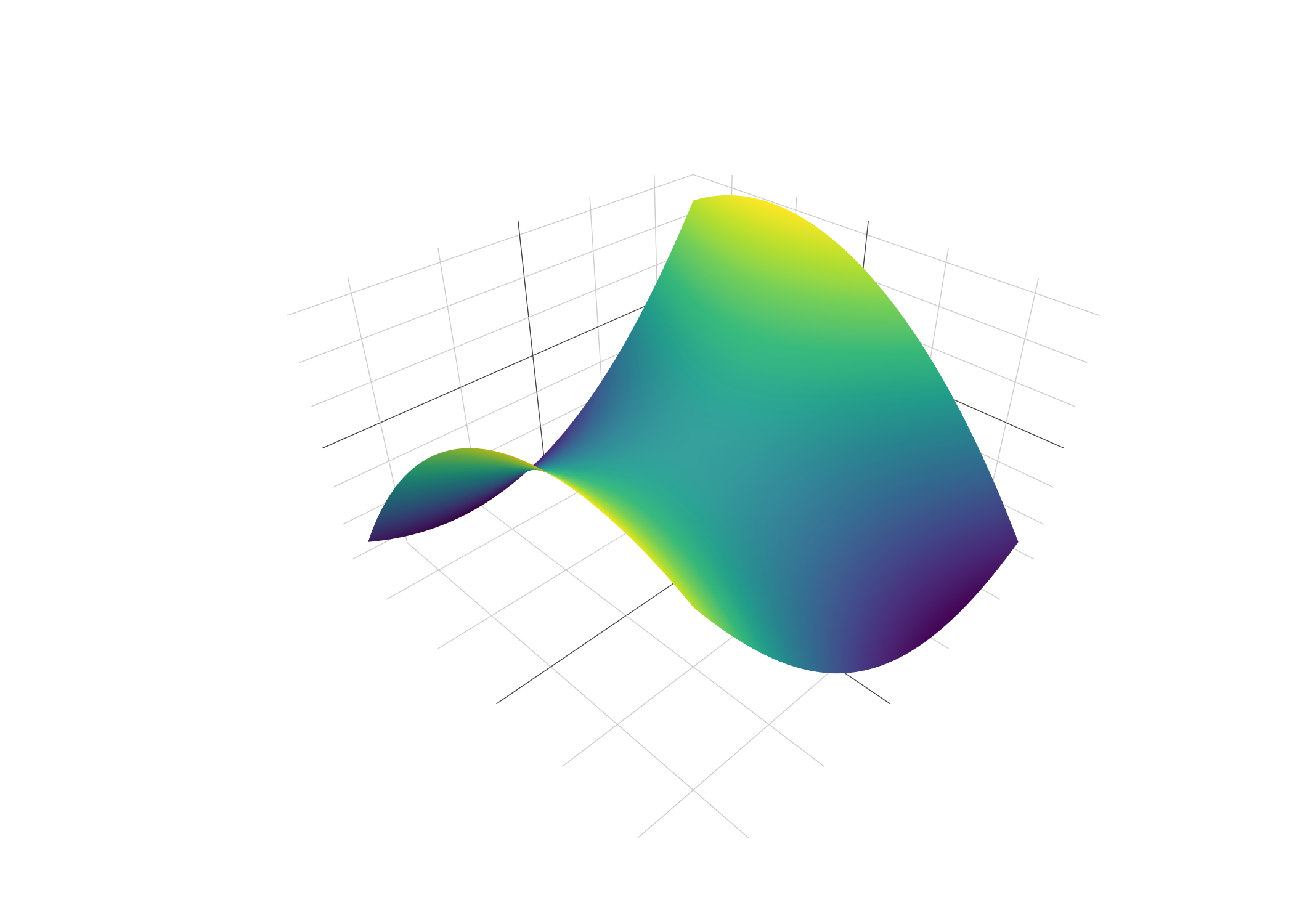}
        \caption{Task 4}
    \end{subfigure}
    \begin{subfigure}[t]{.47\textwidth}
        \centering
        \includegraphics[width = .47\textwidth]{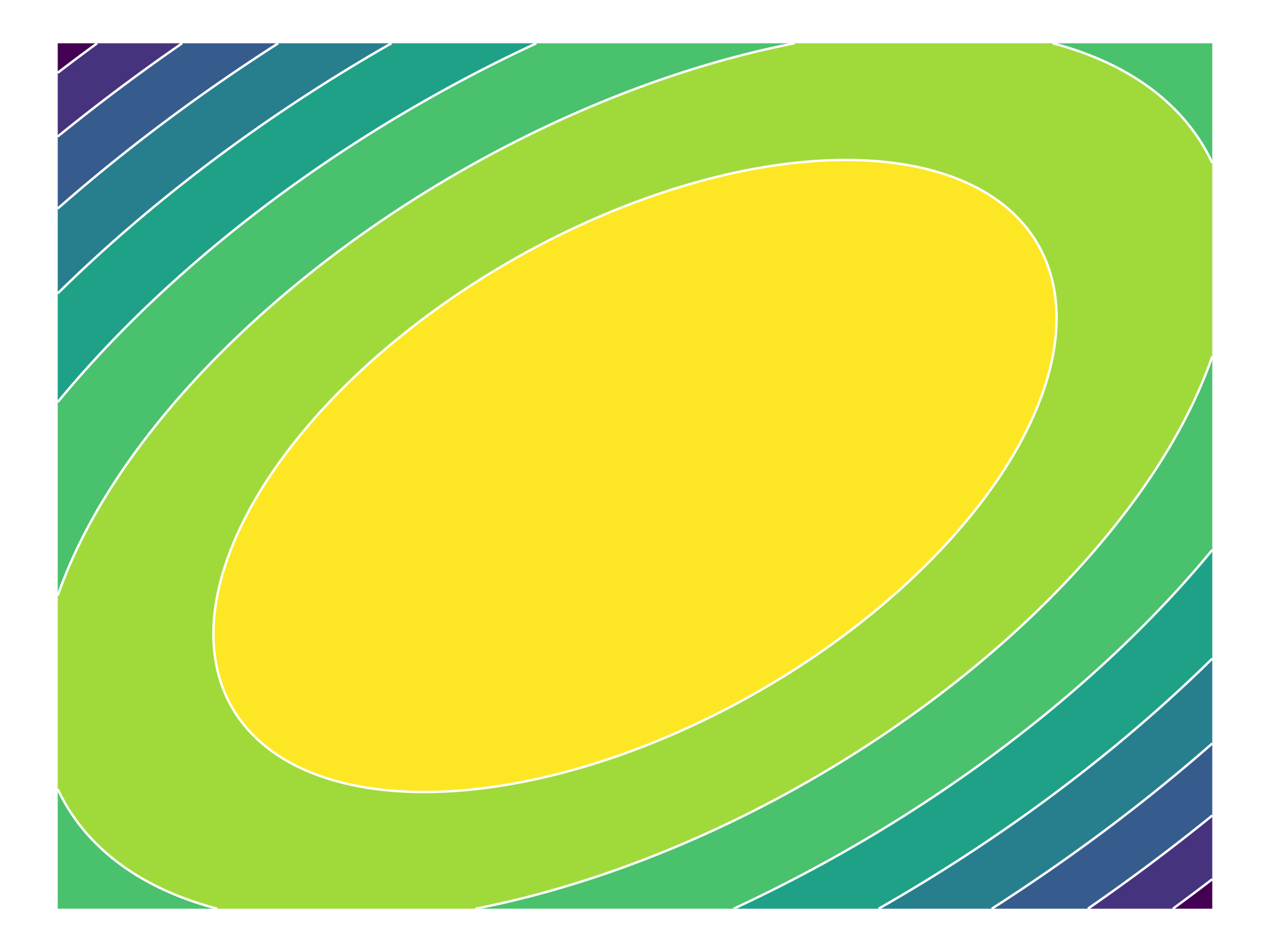}
        \includegraphics[width = .47\textwidth]{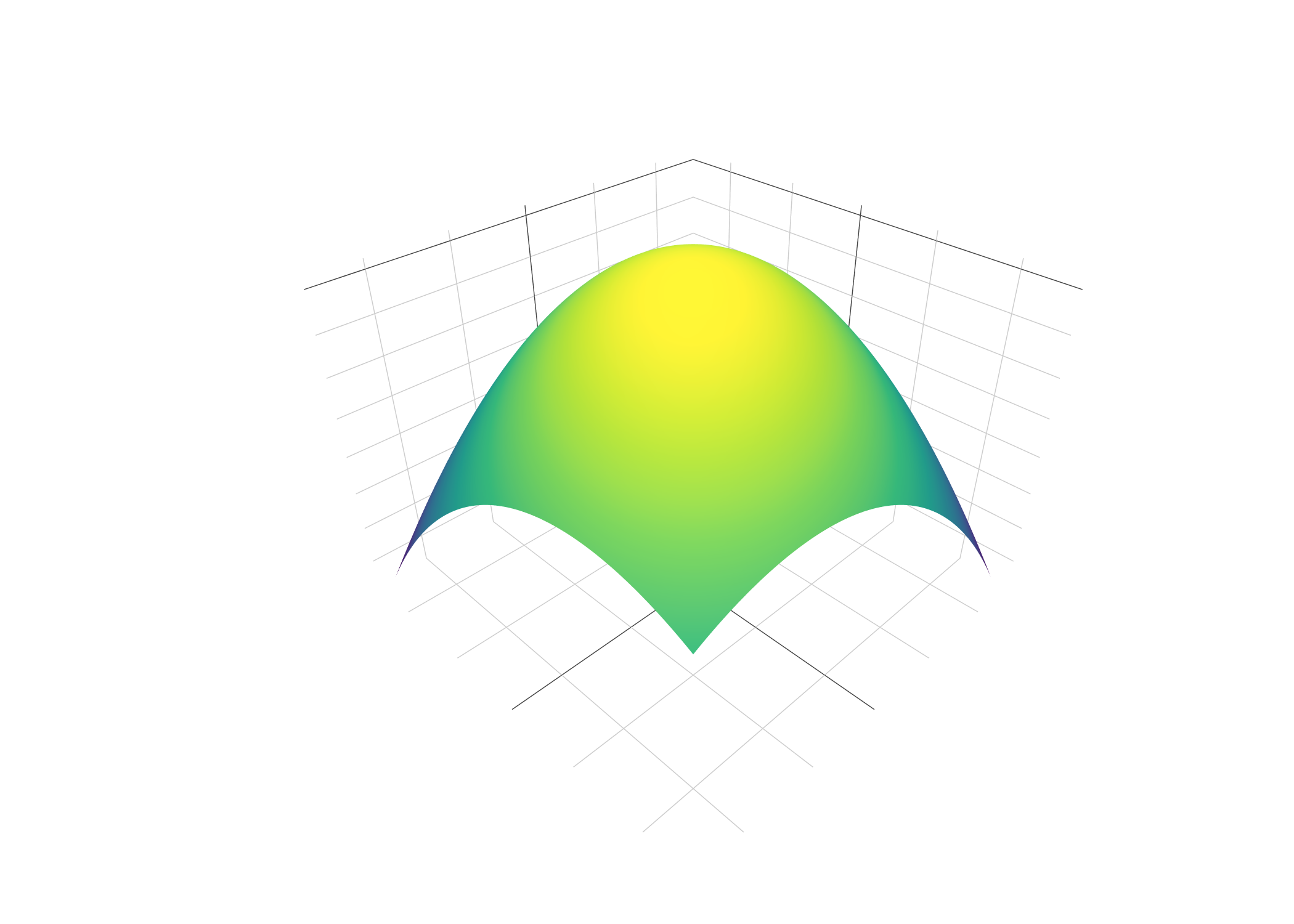}
        \caption{Task 5}
    \end{subfigure}

     \caption{The figures depict the contour plots and surface plot of the predictor variables $X_4$ and $X_5$ for the different tasks.}
     \label{fig:synth1-x4-x5}
\end{figure*}

The second component consists of a quadratic form with respect to variables $X_4$ and $X_5$. The shape of the quadratic forms will vary depending on the values of $a, b, c \in \{-1,1\}$ as illustrated in Table~\ref{table:synth1-params}. The relation between variables~$X_4$ and~$X_5$ is shown in Figure~\ref{fig:synth1-x4-x5}. Although there are tasks that share a similar shape with respect to the projection to one of the variables, the variation in the distribution of variables $X_4$ and $X_5$ across tasks (see Figure~\ref{fig:synth1-density}) will cause the similarity measure to vary.

The predictor variable $X_3$ has no relationship with the outcome. Therefore, if the similarity measure is well-constructed and sufficiently robust, this variable should not affect the values obtained from the measure.

The model used for training was a Random Forest, implemented using the randomForest library of R. To obtain the best model for each task, a grid search strategy was employed, testing combinations of the number of trees (50, 100, 250, and 500) and the number of variables randomly selected at each split (1, 2, 3, 4, and 5). All other hyperparameters were left at their default values. The models were trained on 70\% of the data, with the remaining 30\% used for testing.

The best combination for each model, based on the root mean squared error (RMSE), is represented in Table \ref{table:synth1-rmse}

\begin{table*}
\centering
\caption{RMSE for each task.}
\begin{tabular}{r|c|c|c|c|c|}
\toprule
& Task 1 & Task 2 & Task 3 & Task 4 & Task 5\\
\midrule
RMSE (train) & 0.686 & 0.699 & 0.713 & 0.685&0.717\\
RMSE (test)  & 0.707 & 0.711 & 0.741 & 0.699 & 0.736\\
Number of trees & 250 & 250 & 100 & 250 & 100\\
Features randomly selected & 1 & 3 & 2 & 2 & 4\\
\bottomrule
\end{tabular}
\label{table:synth1-rmse}
\end{table*}

\section{Parkinson dataset}
\label{appendix:parkinson}

The model for the Parkinson dataset was trained in a similar manner to the approach described in in \ref{appendix:dataset-1}. In this case, the number of variables randomly selected at each split was chosen from the values 1, 5, 10, 15, and 19.

\section{Bike-sharing BiciMad dataset}
\label{appendix:bikes}

The dataset is constructed by combining the information available from the Open Data portal of the Empresa Municipal de Transportes de Madrid (EMT) (Municipal Transport Enterprise of Madrid, in English)\footnote{\url{https://opendata.emtmadrid.es/Datos-estaticos/Datos-generales-(1)}} and weather condition data from the Open Data portal of the Madrid City Hall, Spain\footnote{\url{https://datos.madrid.es/portal/site/egob/menuitem.c05c1f754a33a9fbe4b2e4b284f1a5a0/?vgnextoid=fa8357cec5efa610VgnVCM1000001d4a900aRCRD&vgnextchannel=374512b9ace9f310VgnVCM100000171f5a0aRCRD&vgnextfmt=default}}. The predictor variables are shown in Table~\ref{tab:variables}

\begin{table}[htbp]
  \centering
  \footnotesize
  \caption{Description of predictor variables used in the model.}
  \label{tab:variables}
  \begin{tabular}{@{} l c c p{7.1cm} @{}} 
  \toprule
  \textbf{Variable} & \textbf{Type / Range} & \textbf{Unit} & \textbf{Brief description} \\ 
  \midrule
  Lag 1            & numeric             & same as target & Target variable at $t-1$ (one hour earlier). \\
  Lag 2            & numeric             & same as target & Target variable at $t-2$ (two hours earlier). \\
  Lag 3            & numeric             & same as target & Target variable at $t-3$ (three hours earlier). \\
  Lag 4            & numeric             & same as target & Target variable at $t-4$ (four hours earlier). \\
  Lag 5            & numeric             & same as target & Target variable at $t-5$ (five hours earlier). \\[2pt]
  Wind speed       & numeric             & m\,s$^{-1}$    & Mean horizontal wind speed during the current hour. \\
  Wind direction   & numeric (0–360)    & degrees        & Meteorological wind direction (0° = North, 90° = East). \\
  Temperature      & numeric             & °C             & Ambient 2-m air temperature. \\
  Humidity         & numeric (0–100)    & \%             & Relative humidity. \\
  Pressure         & numeric             & hPa            & Surface atmospheric pressure. \\
  Radiation        & numeric             & W\,m$^{-2}$    & Global horizontal solar irradiance. \\
  Precipitation    & numeric             & mm             & Liquid-equivalent precipitation accumulated over the previous hour. \\[2pt]
  Holiday          & binary (0/1)       & —              & Indicator equal to 1 if the date is a national/public holiday. \\
  Month            & integer (1–12)     & —              & Calendar month (1 = January … 12 = December). \\
  Hour             & integer (0–23)     & —              & Hour of day in local (solar) time. \\
  Weekday          & integer (1–7)      & —              & Day of week (1 = Monday … 7 = Sunday). \\ 
  \bottomrule
  \end{tabular}
\end{table}

The proposed neural network is a hybrid-parameter sharing deep learning model with three types of layers, designed for multi-task learning. This architecture leverages shared representations across tasks while preserving task-specific nuances.

The architecture consists of one layer shared across all tasks with the same parameters (hard parameter sharing), followed by task-specific layers with parameters encouraged to be similar by applying a regularization constraint (soft parameter sharing) and (3) fully task-specific layers with independent parameters for each task. The details are as follows: 

\begin{itemize}
    \item \textbf{Input}: A $264\times128\times10$ tensor.
    \item \textbf{Hard parameter sharing}:
    \begin{itemize}
        \item Fully Connected Layer with 264 neurons and ReLU activation.
    \end{itemize}
    \item \textbf{Soft parameter sharing $(\times264)$}:
    \begin{itemize}
        \item Fully Connected Layer with 264 neurons and ReLU activation.
        \item Fully Connected Layer with 128 neurons and ReLU activation.
    \end{itemize}
    \item \textbf{Output Layer $(\times264)$}: A single neuron with ReLU activation (the output must be a positive number).
\end{itemize}

\begin{figure}
\includegraphics[width=\columnwidth]{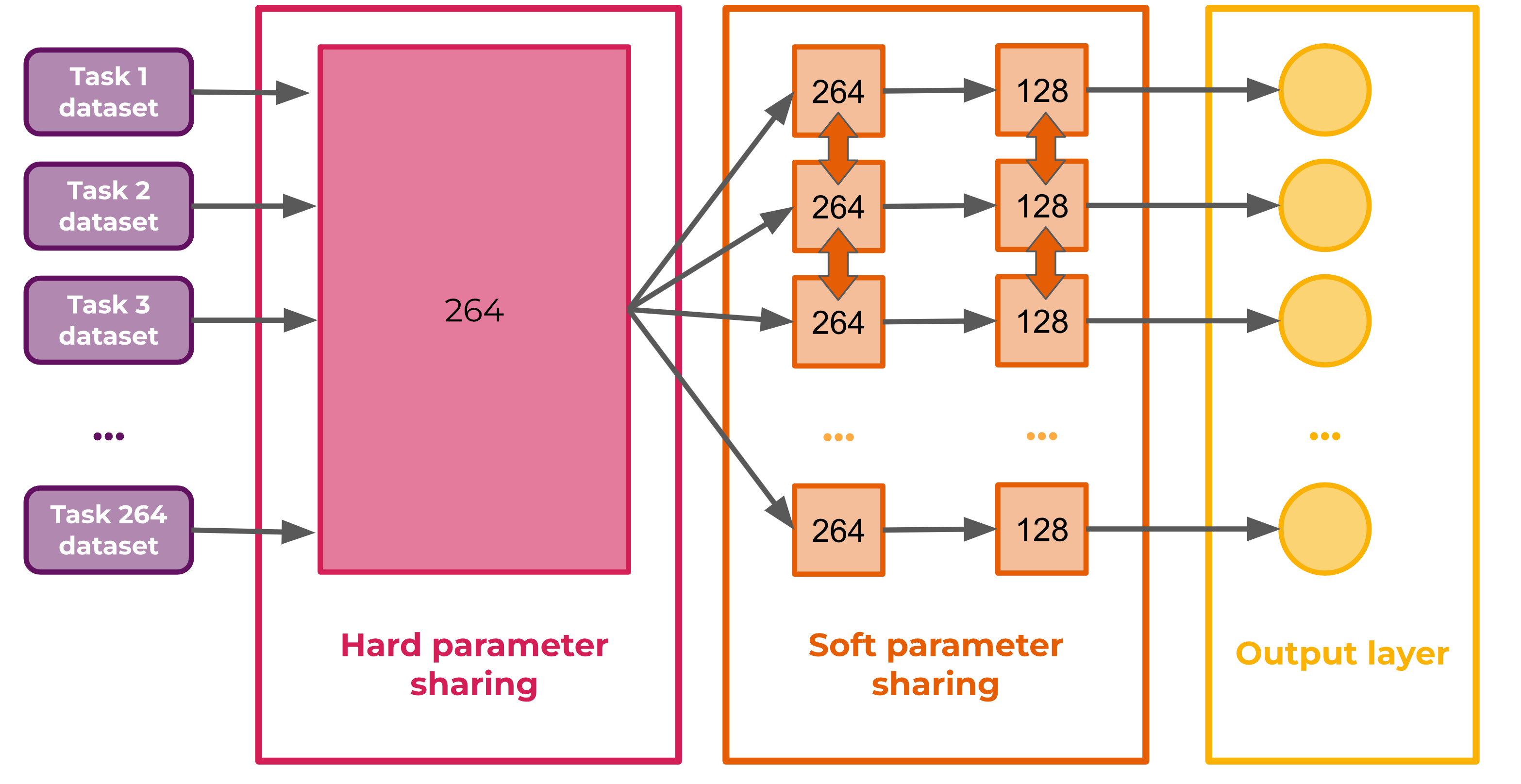}
\centering
\caption{Neural network architecture for Bike-sharing dataset.}
\label{fig:neural-network}
\end{figure}

Figure \ref{fig:neural-network} provides a visual representation of the model.

To obtain the best model, a grid search strategy was employed using a combination of learning rates ($0.1, 0.001, 0.0001,$ and $0.00001$) and a penalty factor values for the regularization constraint in soft parameter sharing ($0.1, 0.01,$ and $0.001$). 

The ALE curves were calculated with a default size of $30$ intervals. However, the variables with fewer than 30 unique values (e.g., the month feature), an appropriate number of intervals was used.

\begin{figure}
\includegraphics[width=.4\columnwidth]{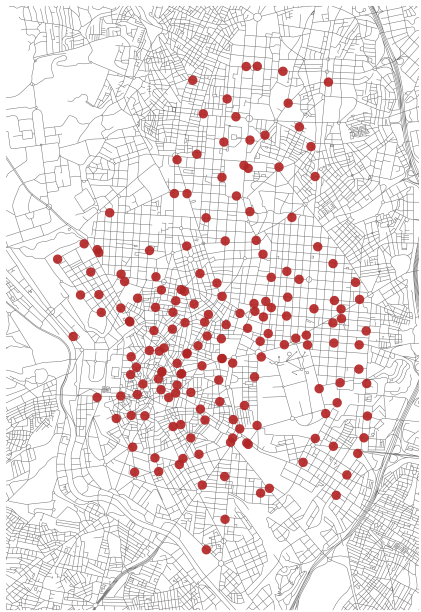}
\centering
\caption{Locations of the stations in the city of Madrid, Spain.}
\label{fig:mapa-estaciones}
\end{figure}

To contextualize the data set, Figure~\ref{fig:mapa-estaciones} displays the locations of the stations Madrid, Spain.

\begin{figure}
\includegraphics[width=\columnwidth]{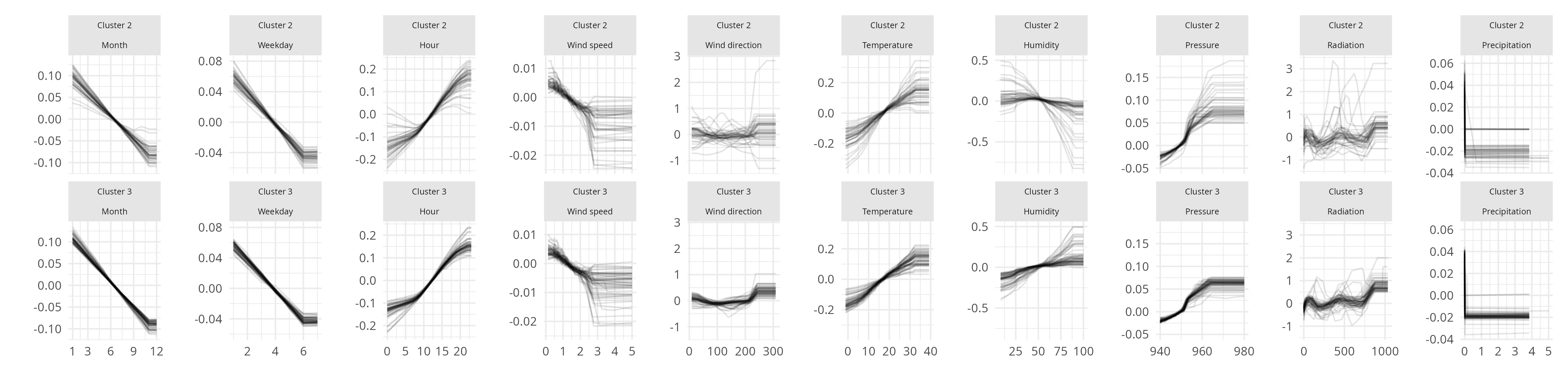}
\centering
\caption{Centered ALE plots for all features from two of the clusters using a Random Forest model.}
\label{fig:clusters-rf}
\end{figure}

The Multi-task Similarity Measure has been computed for the same dataset, this time using a Random Forest model. Figure~\ref{fig:clusters-rf} displays the same two clusters shown in Figure~\ref{fig:ale-clusters-2-3}, but generated with the Random Forest model. It can be observed that the behavior of the Centered ALE plots is very similar.

\subsection{Autoencoder}
\label{appendix:bikes-autoencoder}

The autoencoder consists of a symmetric architecture with an encoder and a decoder. The encoder maps the 16-dimensional input to a 6-dimensional latent representation through two fully connected layers of sizes 16 → 10 → 6, using ReLU activations. The decoder reconstructs the original input by mirroring this structure (6 → 10 → 16), also using ReLU activations, except for the output layer, which uses a linear activation.

The model was trained using the mean squared error (MSE) loss function with the Adam optimizer (learning rate = 0.0001, batch size = 128) for 100 epochs. Early stopping was applied with a patience of 10 epochs.

The complete dataset was used without differentiating between tasks, and the target variable (use) was excluded from the input to the autoencoder.

\section{CelebA dataset}
\label{appendix:celeba}

In this work, we use the CelebA dataset to validate the application of concept bottleneck encoders in a multitask learning setting. CelebA consists of over 200,000 facial images, each annotated with 40 binary attributes representing various semantic features such as facial traits, accessories, and demographic characteristics.

To structure the multitask setting and the concept bottleneck, we split the 40 attributes into two disjoint sets:

\begin{itemize}
\item \textbf{Concept Attributes (used as bottleneck variables):} These 20 attributes are used as interpretable intermediate representations within the concept bottleneck encoder. They represent observable traits that are expected to capture key visual semantics:

\textit{Arched Eyebrows, Bags Under Eyes, Bald, Bangs, Big Lips, Big Nose, Black Hair, Blond Hair, Brown Hair, Bushy Eyebrows, Double Chin, Eyeglasses, Heavy Makeup, High Cheekbones, Mouth Slightly Open, Pale Skin, Receding Hairline, Smiling, Straight Hair, Wavy Hair}.

\item \textbf{Target Task Attributes (used as multitask outputs):} These 20 attributes are predicted from the latent concept representations. Each of them defines a separate output in the multitask learning setup:

\textit{5 o'Clock Shadow, Attractive, Chubby, Goatee, Gray Hair, Male, Mustache, Narrow Eyes, No Beard, Oval Face, Pointy Nose, Rosy Cheeks, Sideburns, Wearing Earrings, Wearing Hat, Wearing Lipstick, Wearing Necklace, Wearing Necktie, Young, Blurry}.
\end{itemize}

All attributes were binarized and normalized prior to training. The dataset was randomly split into training, validation, and test sets following the official partitioning provided by CelebA. Images were resized to $64 \times 64$ pixels, and standard data augmentation techniques (random horizontal flips and normalization) were applied during training.

This setup allows us to evaluate the capacity of the concept bottleneck encoder to extract meaningful latent features that can support multiple downstream tasks while maintaining interpretability.

\end{document}